\documentclass{article}

\usepackage{xcolor}
\usepackage{caption}
\usepackage[most]{tcolorbox}
\usepackage{listings}
\usepackage[T1]{fontenc}
\usepackage[utf8]{inputenc}
\usepackage{hyperref}

\usepackage{amsmath}
\usepackage{amssymb}
\usepackage{mathtools}
\usepackage{amsthm}
\usepackage[capitalize,noabbrev]{cleveref}
\usepackage{array}
\usepackage{multirow}

\usepackage{microtype}
\usepackage{graphicx}
\usepackage{subcaption}
\usepackage{booktabs} 

\usepackage{hyperref}

\usepackage[accepted]{icml2026}

\usepackage{amsmath}
\usepackage{amssymb}
\usepackage{mathtools}
\usepackage{amsthm}

\usepackage[capitalize,noabbrev]{cleveref}

\theoremstyle{plain}

\theoremstyle{definition}

\theoremstyle{remark}

\usepackage[textsize=tiny]{todonotes}
\usepackage[]{color-edits}
\addauthor{vc}{red}
\addauthor{wc}{blue}
\addauthor{sk}{olive}
\addauthor{aw}{violet}
\addauthor{cd}{magenta}
\addauthor{rc}{yellow}
\addauthor{sy}{teal}

\newcommand{\framework}[1]{\texttt{#1}}
\newcommand{\model}[1]{\texttt{#1}}
\newcommand{\taskcount}{154}
\newcommand{\variantcount}{179}
\newcommand{\fullcount}{333}

\icmltitlerunning{GameDevBench: Evaluating Agentic Capabilities Through Game Development}

\begin{document}
\title{\textbf{GameDevBench: Evaluating Agentic Capabilities Through Game Development}}

\twocolumn[
  \icmltitle{GameDevBench: Evaluating Agentic Capabilities Through Game Development}



    \icmlsetsymbol{equal}{*}

    \begin{icmlauthorlist}
      \icmlauthor{Wayne Chi}{cmu}
      \icmlauthor{Yixiong Fang}{cmu}
      \icmlauthor{Arnav Yayavaram}{cmu}
      \icmlauthor{Siddharth Yayavaram}{cmu}
      \icmlauthor{Seth Karten}{princeton}
      \icmlauthor{Qiuhong Anna Wei}{cmu}
      \icmlauthor{Runkun Chen}{cmu}
      \icmlauthor{Alexander Wang}{cmu}
      \icmlauthor{Valerie Chen}{cmu}
      \icmlauthor{Ameet Talwalkar}{cmu}
      \icmlauthor{Chris Donahue}{cmu}
    \end{icmlauthorlist}
    
    \icmlaffiliation{cmu}{Carnegie Mellon University, Pittsburgh, PA, USA}
    \icmlaffiliation{princeton}{Princeton University, Princeton, NJ, USA}
    
    \icmlcorrespondingauthor{Wayne Chi}{waynechi@andrew.cmu.edu}
    
    \icmlkeywords{Machine Learning, ICML}
    
    \vskip 0.3in
]



\printAffiliationsAndNotice{}  

\begin{abstract}
Despite rapid progress on coding agents, progress on their multimodal counterparts has lagged behind.
A key challenge is the scarcity of evaluation testbeds that combine the complexity of software development with the need for deep multimodal understanding.
In game development, agents must navigate large, dense codebases while manipulating intrinsically multimodal assets such as shaders, sprites, and animations within a visual game scene.
We present \texttt{GameDevBench}, the first benchmark for evaluating agents on game development tasks.
\texttt{GameDevBench} consists of \fullcount~tasks derived from web and video tutorials. 
Tasks require significant multimodal understanding and are complex---the average solution requires over three times the lines of code and file changes compared to prior software development benchmarks. 
Agents struggle with game development, with the best agent and method solving only $53.8\%$ of tasks.
We find a strong correlation between perceived task difficulty and multimodal complexity, with average success rate dropping from $51.4\%$ on gameplay-oriented tasks to $33.0\%$ on 2D graphics tasks.
To improve multimodal capability, we introduce two simple image and video-based feedback mechanisms for agents.
Despite their simplicity, these methods consistently improve performance, increasing \model{GPT-5.4}'s performance from $41.1\%$ to $52.0\%$ when given visual feedback.
We release our code at 
\url{https://github.com/waynchi/gamedevbench}




\end{abstract}

\section{Introduction}

\begin{figure*}[t!]
    \centering
    \includegraphics[width=0.865\linewidth]{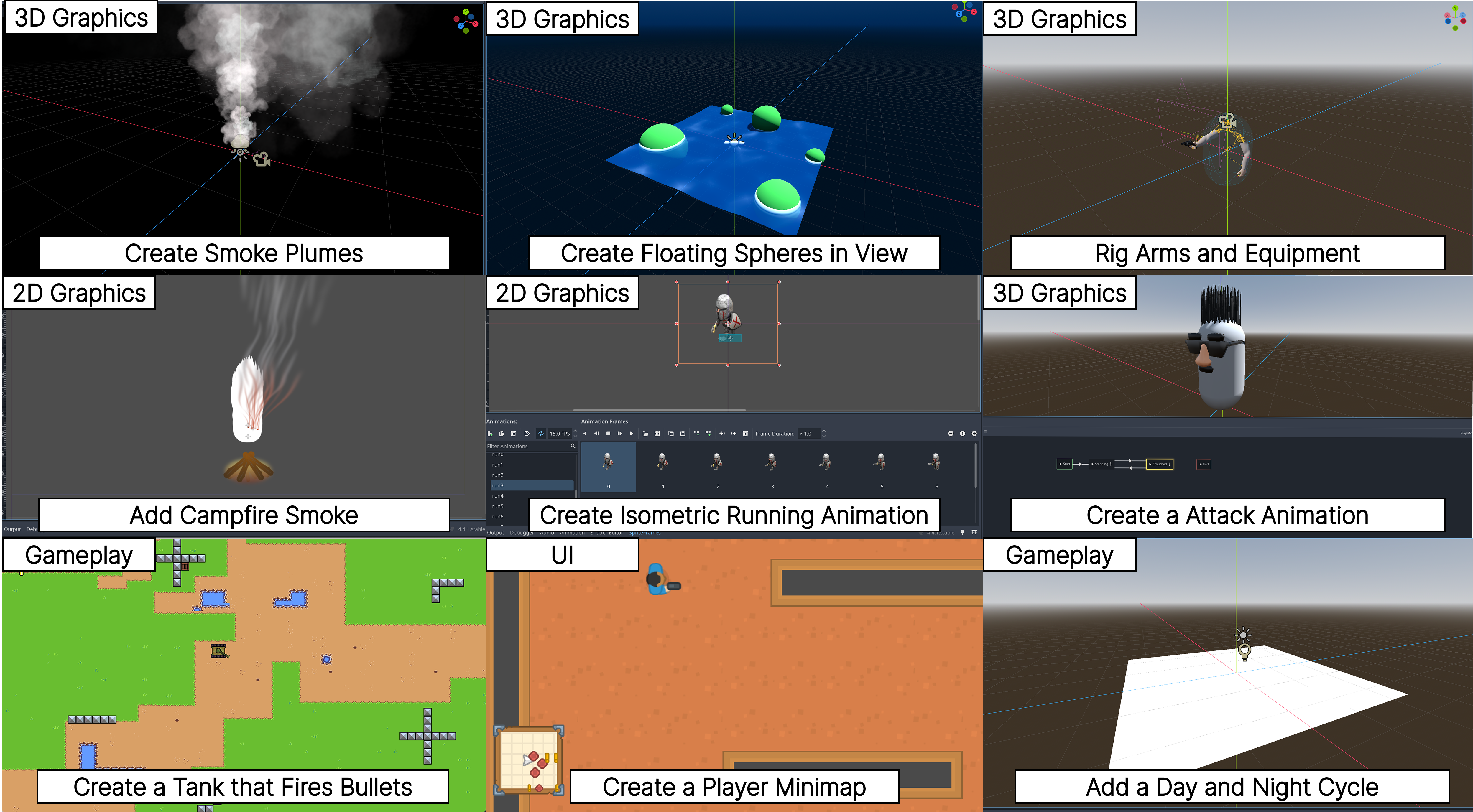}
    \caption{We present \texttt{GameDevBench}, a benchmark for evaluating an agent's ability to solve complex and multimodal game development tasks in a modern game engine.}
    \label{fig:multiple-examples}
\end{figure*}

Progress on multimodal language model (LM) agents has lagged behind that of their unimodal counterparts~\citep{yang2024swebenchmultimodalaisystems, jimenez2024swebenchlanguagemodelsresolve, zhou2024webarenarealisticwebenvironment, koh2024visualwebarenaevaluatingmultimodalagents}.
Agentic game development---despite its inherent multi-modality, increasing public interest, and a rich history combining artificial intelligence and games~\citep{Vinyals2019GrandmasterLIA, Schrittwieser2019MasteringAGA, Silver2018AGRA, silver2016mastering, jagli2024artificial, filipovic2023role, yakan2022analysis}---has largely been overlooked by the research community.
Most prior works focus on specific goals within game development such as next frame prediction~\citep{Valevski2024DiffusionMAA, Oh2015ActionConditionalVPA}, which replaces the graphics engine, procedural content generation~\citep{summerville2018proceduralcontentgenerationmachine, shaker2016procedural}, which replaces asset creation, or game playing agents~\citep{Vinyals2019GrandmasterLIA, silver2016mastering}, which replaces the non-player characters (NPCs) and opponents. 
There has been little to no research on agentic use for general game development (i.e., developing games within a game engine), most likely because it seemed inconceivable until recently.
As LM agent capabilities continue to improve, it seems natural to ask: 
can agents develop video games? 

Game development combines many desirable characteristics for a challenging benchmark in a modern agentic domain.
First, tasks are \textbf{complex and context-rich} with projects often spanning large amounts of files, assets, and folders akin to that of traditional software development~\citep{yang2024swebenchmultimodalaisystems}. 
Second, tasks are inherently \textbf{multimodal}, requiring visual understanding of both static elements (e.g., map or scene layouts) and temporal dynamics (e.g., animations or movement) to accurately assess project state.
Lastly, task solutions are \textbf{deterministically verifiable} through code which alleviates the need for approaches such as LLM-as-a-Judge~\citep{zheng2023judgingllmasajudgemtbenchchatbot} which are often subject to biases~\citep{wang2023largelanguagemodelsfair, koo2024benchmarkingcognitivebiaseslarge}.
For example, it is possible to verify that the correct animation was used by checking animation states at each frame.
This combination of features makes game development an ideal environment to evaluate complex, multi-modal agentic capabilities. 

In this work, we study an agent's ability to solve complex game development tasks for a modern game engine. 
To our knowledge, this is the first work evaluating this capability.
Game development typically involves creating and editing artifacts such as sprite sheet animations, collision shapes, game logic scripts, and scene layouts in a GUI (Graphical User Interface) called the game editor.
A game engine then processes these artifacts into a runnable game build. 
Common examples of game engines include Unity, Unreal Engine, and Godot, each of which provides both an editor and an engine.
Game development tasks are deceptively complex.
For example, the ``simple'' task of creating an Italian plumber for a platformer game would require creating animations for various states such as idling, jumping, or running, setting up a collider to allow for jumping on enemies such as turtles, writing scripts to allow for control, adding sound effects for actions, and more.

We focus our work on the Godot environment for several reasons.
First, Godot is fully open sourced under the MIT license which makes it easy to extend and release alongside the benchmark.
Second, Godot is an increasingly popular game development engine, with 770 and 1185 releases on Steam in 2024 and 2025.
Third, Godot's environment strongly resembles Unity, which is by far the most popular game development engine.
Lastly, Godot projects (not including assets such as images) can be represented in code which makes it simple to extend existing LLM agent capabilities without having to construct specific tool-use APIs.

We present \texttt{GameDevBench}, the first benchmark for evaluating an agent's ability to solve game development tasks.
Tasks are created by analyzing and processing Godot YouTube and web tutorials.
These tutorials span a wide range of topics such as 2D sprite animations, character controllers (i.e., character movement), colliders and platforms, shader usage, particle effects, among others.
This ensures that tasks are not only diverse, but also align with common game development needs.
Tasks are incredibly complex and content-rich.
Not only do they require a deep understanding of various file types and assets (e.g., images), tasks on average require more than \textit{three times} the number of lines of code changes compared to SWE-Bench~\citep{yang2024swebenchmultimodalaisystems}.
For each task, agents are given a project folder with code and various assets, as well as an instruction as is standard in software benchmarks~\citep{yang2024swebenchmultimodalaisystems}.
Task success is evaluated using tests built within Godot's scripting framework.
This allows us to deterministically test for features such as physics or polygonal shapes.
Additionally, each task comes with a verified reference solution.
All code and task project files for \texttt{GameDevBench} are released publicly.

We found that while agents are increasingly capable, they still struggle with the majority of game development tasks.
With additional multimodal support, the best agent succeeds at only $53.8\%$ of the tasks.
In particular, models perform significantly worse when the tasks require increased multimodal understanding. 
For example, agents perform nearly fifty percent as well on gameplay-oriented tasks compared to 2D graphics tasks ($51.4\%$ vs $33.0\%$ averaged across all evaluated agents).

To improve agent multimodal capabilities, we propose two methods that provide agents with multi-modal feedback when solving a task.
One method provides a screenshot view of the editor's current state via a Model Context Protocol (MCP) server~\citep{anthropic2024modelcontext} while another records a video of the game scene.
Despite their simplicity, we found that both methods are effective empirically, increasing agent performance across almost all models.

\section{Benchmark Construction}\label{sec:benchmark-construction}

\begin{figure*}[th!]
    \centering
    \includegraphics[width=0.8\linewidth]{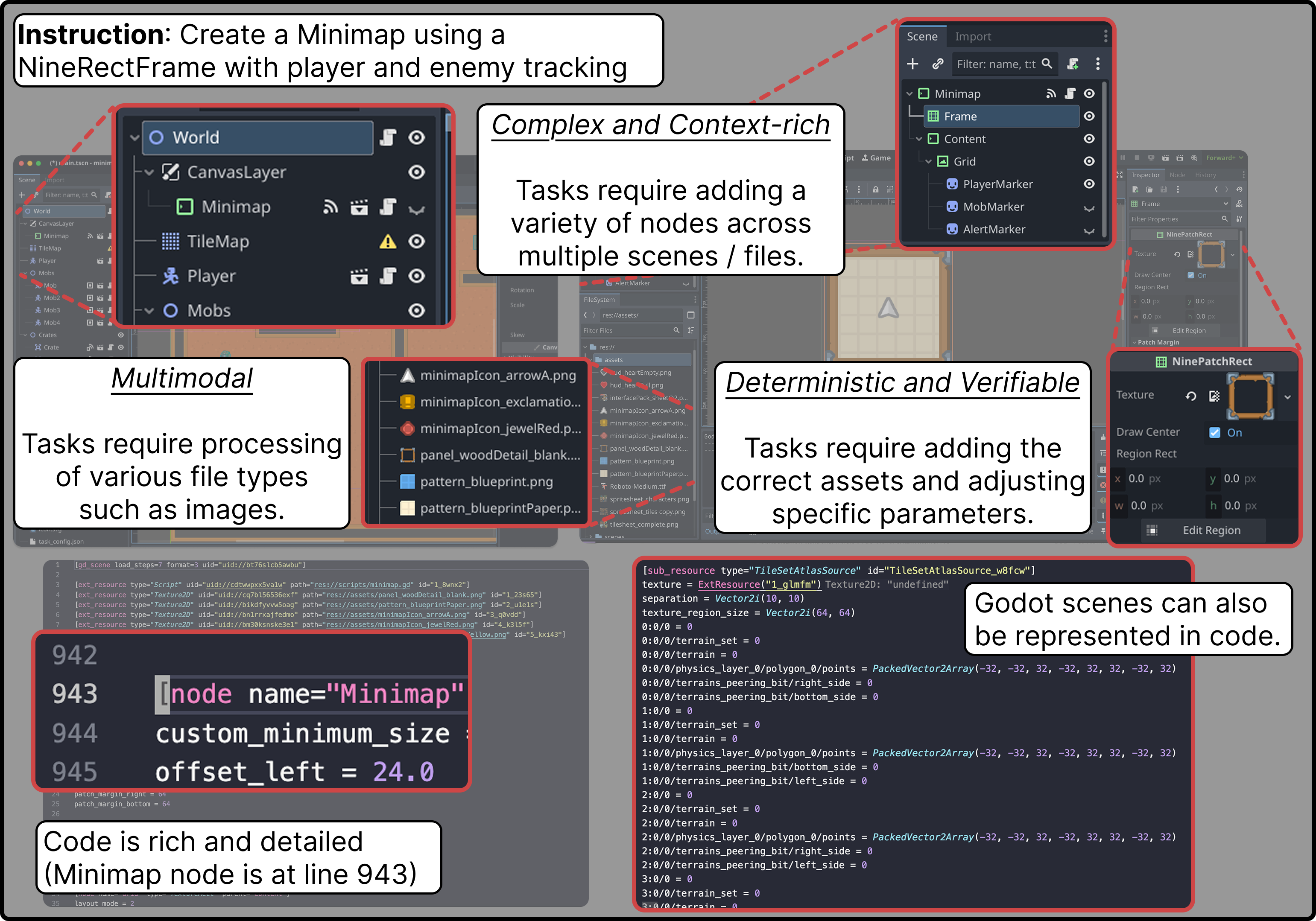}
    \caption{This is an example task from \texttt{GameDevBench} that requests for the creation of a UI minimap.
    Top is the visual GUI representation and highlighted points of interest.
    Bottom is the same scenes and files represented in code.
    Tasks can be solved via the editor or entirely through code although either method requires understanding multimodal assets.
    Game development tasks are complex and require editing dense files, identifying and visually understanding  various assets, and navigating various nodes (game elements) and scenes (a collection of nodes). 
    }
    \label{fig:single-example}
\end{figure*}

\texttt{GameDevBench} consists of game-development tasks distilled from online tutorials (e.g., ``add a walking animation using the given spritesheet''). 
GitHub repositories are a rich source of data, but can be noisy and poorly documented.
Additionally, unlike prior work~\citep{jimenez2024swebenchlanguagemodelsresolve} on benchmarking general software development, there are no obvious popular open source game repositories to choose from.
The game development community, however, has created an abundance of online tutorials---many of which come with solution repositories---that guide developers through common development use cases.
We use a multi-step pipeline to construct game development tasks using these tutorials.

\subsection{Stage 1: Data Preparation}

Game development tutorials primarily come in either text or video formats.
For all tutorials, we search for and filter to only include Godot 4 tutorials that include a corresponding GitHub repository with permissive, open-source licenses.

\textbf{Video.}
We source our video tutorials from YouTube.
To convert video into text, we use a popular YouTube transcription API\footnote{https://github.com/jdepoix/youtube-transcript-api} to extract the text transcript from each video.
To search for a matching GitHub repository, we parse the video description for any GitHub links.
The final result is a folder for each tutorial containing the transcript and the corresponding GitHub repository.
We process 102 video tutorials which were selected based on view count.
Each tutorial averages 29 minutes of content. 
In the end, we use 57 tutorials as not all tutorials are usable due to non-functional repositories or mislabeled Godot versioning.

\textbf{Web.}
For web text tutorials, we source from ``Godot Recipes by KidsCanCode''~\citep{kidscancode_godot_recipes_4x}, which is listed on the community resources page in Godot's official documentation. 
We scrape the webpages using a Python script with the goal of mirroring the structure of processed video tutorial folders. 
The end result is 99 tutorial folders, each of which contains the tutorial text content, a media directory of visual data downloaded from the webpage, a GitHub repository, as well as a metadata JSON containing information such as the tutorial URL.
Finally, we ask an LLM to sort tutorials based on suitability for task creation and use the top 31 tutorials for subsequent task construction. 


\subsection{Stage 2: Automatic Task Construction}
Given the tutorial folder, the agent is asked to create tasks where 1) instructions adhere to the tutorial, 2) task files are created directly based on existing files in the repository, and 3) unit tests must only test for features explicitly requested in the instructions.
Access to the solution repository is crucial as it allows the agent to create tasks that it would not normally have the capability to solve or create.
At the agent's discretion, each tutorial is split into multiple tasks to capture more well-defined skills.
For example, the agent could decompose a platformer tutorial into tasks on character animation, controls and colliders, and tilemap construction.
We use the \framework{Codex} Agent with the GPT-5 family of models to construct tasks from each tutorial.
\framework{Codex} was chosen primarily due to its API limits and availability at the time; 
we did not notice any significant differences between agents such as \framework{Claude Code}.
We create 202 initial tasks with an average of 1.3 tasks per tutorial.
The full prompt can be found in Appendix \ref{appendix:task_construction}. 

\subsection{Stage 3: Task Refinement}
After stage 2, we found the majority of tasks to be sensible at a high level (i.e., task instructions were reasonable and matched the tutorial).
However, similar to prior work~\citep{chi2025editbenchevaluatingllmabilities}, the agent was not able to perfectly create tasks and tests.
We conducted a preliminary study on a small subset of 41 tasks where human annotators reviewed tasks and documented any issues observed.
The study found that 43\% of tasks were issue free, 50\% of tasks had issues that required minor updates such as scenes being off-camera, tests asserting for non-existent instructions, or accidental references to other portions of the tutorial, and 7\% of tasks contained major issues that made them difficult to fix.
Since most of the errors were minor and easily caught, we employed a hybrid process to refine tasks. 
Based on the preliminary study, we constructed prompts and checklists to catch the most common mistakes.
We then employed an agent to automatically verify and fix those mistakes based on the checklists.
We re-use this prompt (Appendix \ref{appendix:task_refinement}) when processing all future tutorials and tasks.

\subsection{Stage 4: Human Annotation.}
Lastly, 8 human annotators, 5 of whom have prior game development experience, reviewed all tasks.
Annotation served three goals.
First is to ensure that tasks are verified for correctness and resolvability as is common practice~\citep{yang2024swebenchmultimodalaisystems, chi2025editbenchevaluatingllmabilities}.
Annotators are instructed to look for and fix any ambiguous instructions, conflicting instructions, and overly strict tests.
Additionally, annotators are asked to mark and remove any tasks that had any other issues.
Second, we ask annotators to create variations of existing tasks similar to prior work~\citep{zhou2024webarenarealisticwebenvironment}.
An example would be two tasks that differ based on the requested animation used in a spritesheet (e.g, selecting the walking vs running animation frames).
In total, we create \taskcount~tasks and \variantcount~task variants.
Lastly, we asked annotators to annotate whether tasks were considered \texttt{easy} or \texttt{hard} based on the task's required multimodal understanding.
In total, we create 131 \texttt{easy} and 202 \texttt{hard} tasks.
Annotation instructions can be found in Appendix \ref{appendix:human_annotation}.

\section{GameDevBench}

Game development sits at the intersection of creative expression and software development.
As such, \texttt{GameDevBench} features a diverse set of tasks that are inherently multi-modal, complex and context-rich.

\subsection{Task Categories}
To our knowledge, there is no existing taxonomy of game development tasks performed within a game editor or game engine.
To better understand our task diversity and enable deeper analysis, we categorize each task along two axes.

\textbf{Categorization by skill set.}
We induce a task categorization based on the underlying game development skills required by each task (Table~\ref{tab:godot-domains}).
Specifically, we adopt a bottom-up categorization procedure: 
we first obtain fine-grained skill annotations for each task by asking \model{GPT-5-mini} to label each task.
Labels are then abstracted into higher-level skill categories through a separate request to \model{GPT-5-mini}.
These categories are subsequently reviewed and refined by game developers to ensure consistency and validity.
This process yields four skill categories: 2D graphics and animation, 3D graphics and animation, gameplay logic, and user interface.


\begin{table*}[t!]
\centering
\small
\caption{Skill categories for Godot-related development tasks.}
\label{tab:godot-domains}
\begin{tabular}{@{}m{0.15\linewidth}m{0.35\linewidth}m{0.3\linewidth}>{\raggedleft\arraybackslash}m{0.1\linewidth}@{}}
\toprule
\textbf{Skill} & \textbf{Definition} & \textbf{Examples} & \textbf{\% Tasks} \\
\midrule

2D Graphics \newline
and Animation &
Tasks involving the creation, rendering, and animation of two-dimensional visual content. &
Sprite animation, TileMap setup, 2D shader effects &
33.3\% \\
\addlinespace

3D Graphics \newline
and Animation &
Tasks involving the construction, rendering, and animation of three-dimensional scenes. &
Material tuning, Skeletal animation, Camera rigs &
26.7\% \\
\addlinespace

User Interface &
Tasks concerning the design and implementation of interactive user interfaces. &
HUD layout, Menu navigation, UI theming &
20.1\% \\
\addlinespace

Gameplay Logic &
Tasks focused on implementing game rules and behaviors such as motion and collisions. &
Enemy AI states, Signal-driven events, Collision detectors, Character controllers &
19.8\%\\
\addlinespace
\bottomrule
\end{tabular}
\end{table*}

\begin{figure}[h!]
    \centering
    \includegraphics[width=0.95\linewidth]{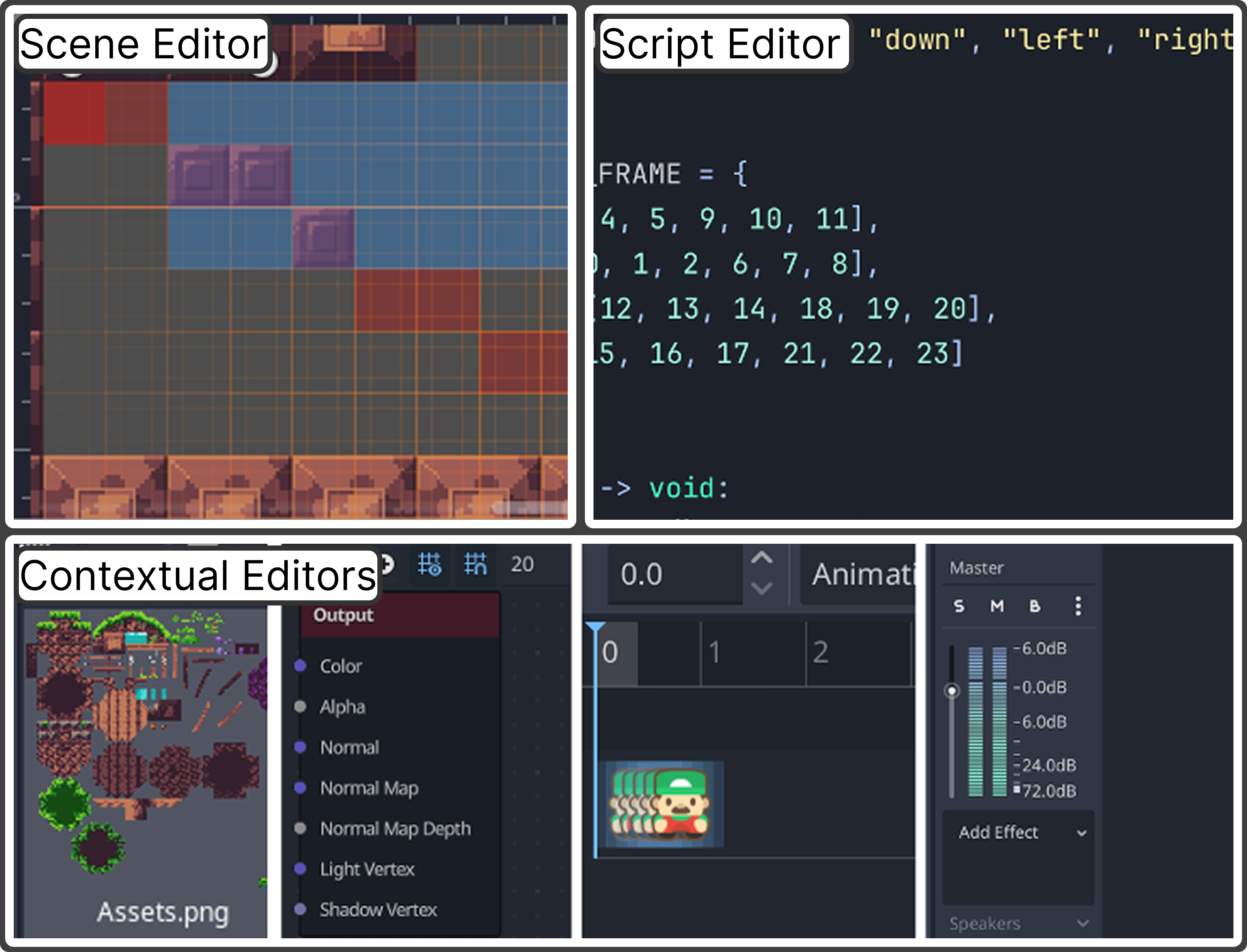}
    \caption{Types of editors in Godot. 
    Top-left is the scene editor.
    Top-right is the script editor.
    The bottom contains various contextual editors.
    From left to right: tilemap, shader, animation, and audio editors.
    Contextual editors surface depending on use case.
    Typically, tasks that use contextual editors require deeper multi-modal understanding.
    }
    \label{fig:editor}
\end{figure}

\textbf{Categorization by editor type.}
Godot contains several different types of editors that users use to resolve various tasks.
There are three main types of editors within Godot.
The scene editor (Figure~\ref{fig:editor}, top-left) allows the user to modify the game scene by constructing level maps or placing and editing objects.
The script editor (Figure~\ref{fig:editor}, top-right) is a built-in code editor.
Contextual editors appear on the bottom panel depending on what the user is editing (Figure~\ref{fig:editor}, bottom).
For example, when editing an animation resource, the animation editor will appear.
Some of the most common contextual editors include the animation, audio, shader, and tileset editors.
We categorize each task in the benchmark by asking \model{GPT-5-mini} to determine the editors that a user would need to solve the task.
While the agent may not directly interact with these editors, the type of editor a user would use provides a strong proxy for task categorization.

For simplicity, we assign each task one skill category and one editor type. 
We provide multiple examples with their skill category and editor type in Appendix~\ref{appendix:task-example}.

%

\subsection{Features of \texttt{GameDevBench}}

\texttt{GameDevBench} has a unique set of features which we describe as follows. We provide additional task statistics in Appendix~\ref{appendix:statistics}.

\begin{figure*}[t!]
    \centering
    \begin{tabular}[t]{@{}c@{\hspace{0.0\linewidth}}c@{\hspace{0.0\linewidth}}c@{}}
        \begin{minipage}[c]{0.36\linewidth}
            \centering
            \includegraphics[width=\linewidth]{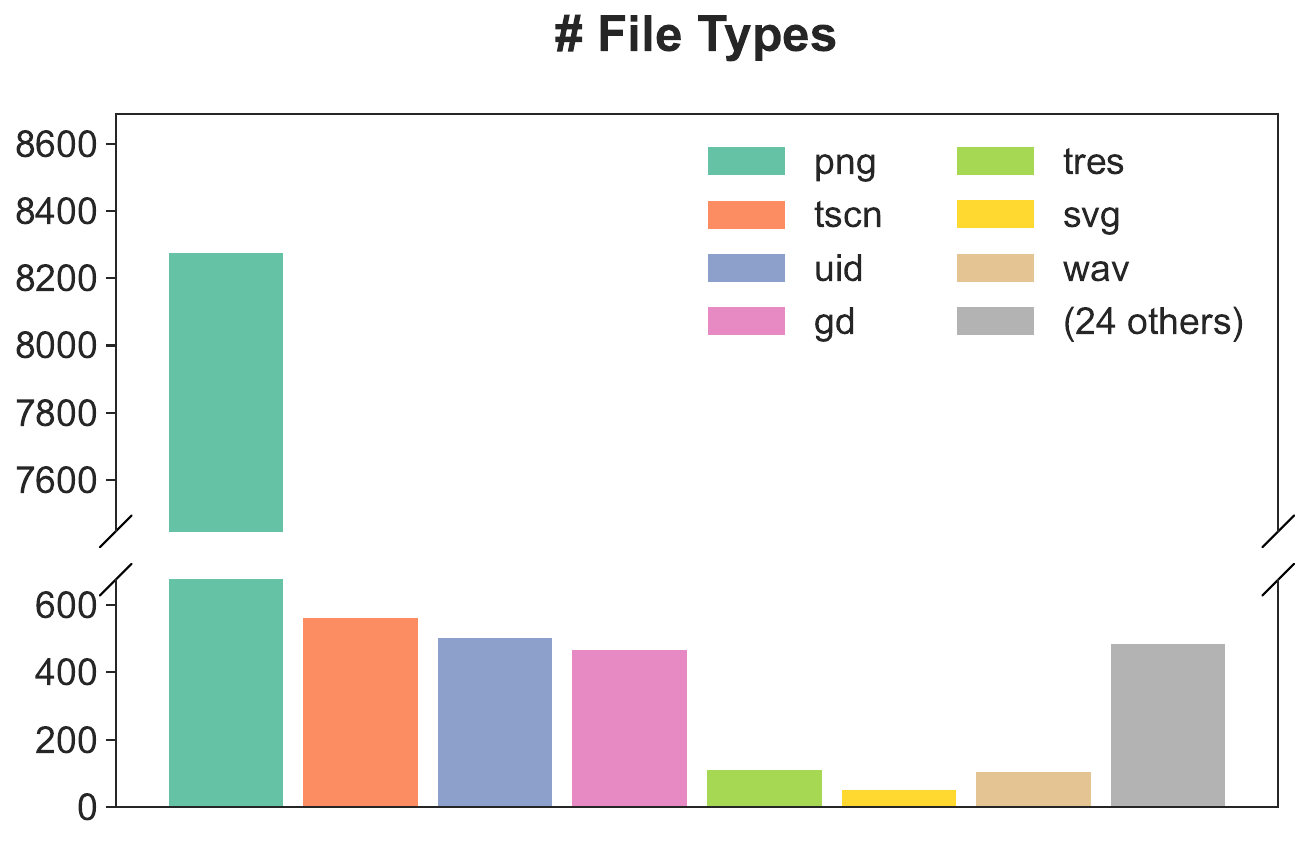}
        \end{minipage}
        &
        \begin{minipage}[c]{0.29\linewidth}
            \centering
            \includegraphics[width=\linewidth]{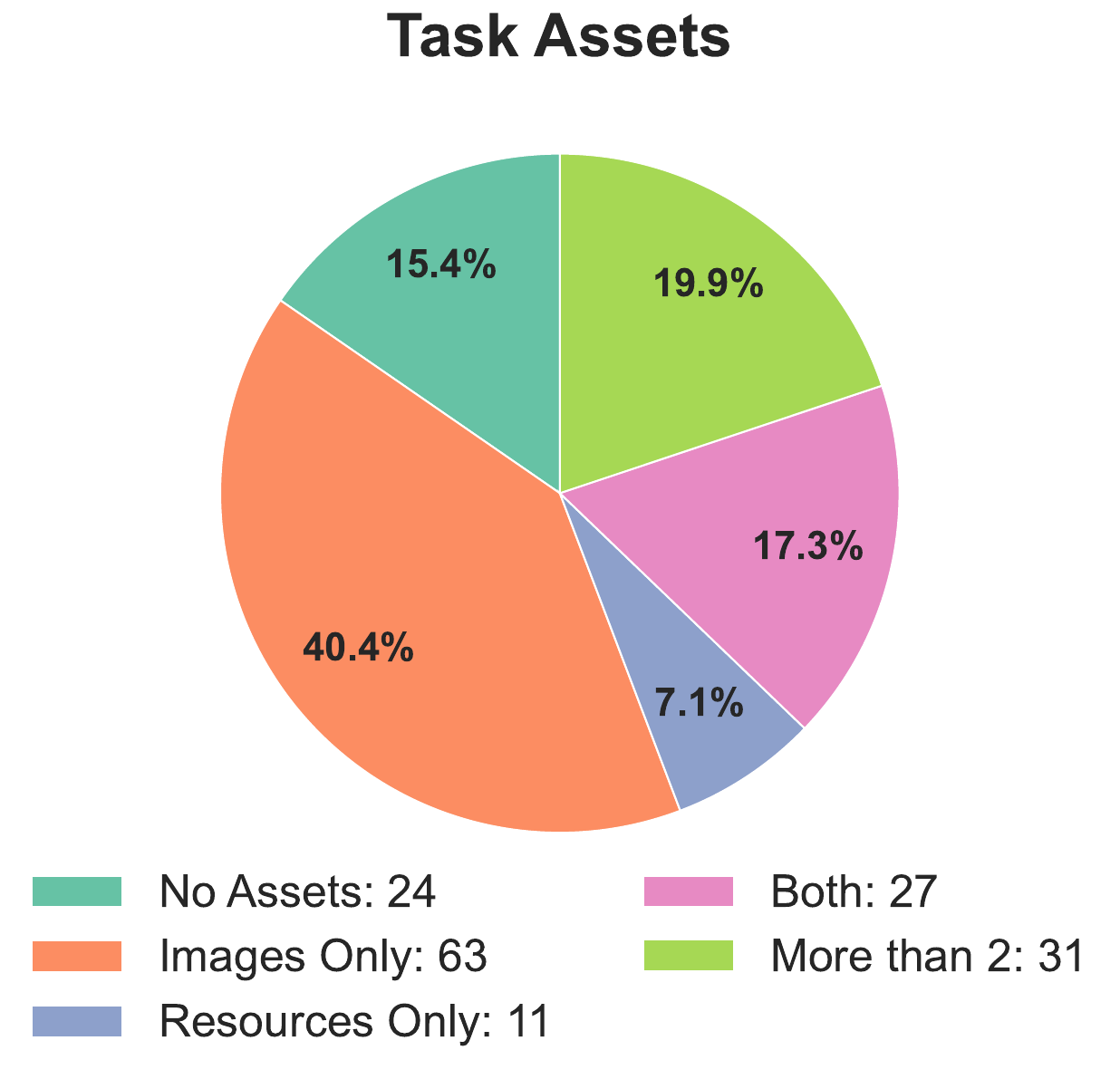}
        \end{minipage}
        &
        \begin{minipage}[c]{0.35\linewidth}
            \centering
            \resizebox{\linewidth}{!}{%
            \begin{tabular}{llrr}
            \toprule
             & & \textbf{Mean} & \textbf{Max} \\
            \midrule
            \multirow{4}{*}{Overview} & \# Files & 67.7 & 1929 \\
             & \# Filetypes & 6.7 & 18 \\
             & \# Lines of Code & 518.9 & 20072 \\
             & \# Nodes & 19.7 & 982 \\
            \midrule
            \multirow{4}{*}{Gold Patch} & \# Files Edited & 4.7 & 17 \\
             & \# Filetypes Edited & 3.2 & 6 \\
             & \# Lines Edited & 114.1 & 1949 \\
             & \# Nodes Edited & 2.1 & 22 \\
            \midrule
            \multirow{2}{*}{Images} & \# Images & 53.4 & 1920 \\
            & Image Size (px) & 119.5K & 16.8M \\
            \bottomrule
            \end{tabular}
            }
        \end{minipage}
    \end{tabular}
    \caption{\texttt{GameDevBench} features a diverse amount of filetypes (31 different types, left). 
    The vast majority of tasks contain either images, resources (e.g., Shaders), or multiple asset types (middle).
    Each task contains multiple scripts and scenes, both of which are context-rich and require a significant amount of tokens to process (right).}
    \label{fig:statistics}
\end{figure*}

\textbf{Diverse file types across tasks.}
Unlike agentic benchmarks in the software domain~\citep{jimenez2024swebenchlanguagemodelsresolve}, \texttt{GameDevBench} requires that agents handle a wide variety of filetypes across various modalities (Figure~\ref{fig:statistics}, left).
In fact, the vast majority of tasks ($82.4\%$) contain additional assets such as images (.png), text fonts (.ttf), shaders (.gdshader), audio (.wav), and other asset resources (.tres).
As such, \texttt{GameDevBench} inherently tests the multimodal capabilities of agents.

\textbf{Diverse task types.}
While there are other domains (such as frontend development~\citep{Zhu2025FrontendBenchABA,Si2024Design2CodeHFA} or slide generation~\citep{liang2025slidegencollaborativemultimodalagents}) that intersect multimodality and code generation, most of these domains focus on tasks similar to user interface generation.
On the other hand, \texttt{GameDevBench} features a diverse task set.
Across the \fullcount~benchmark tasks, domains are distributed as follows:
(33.3\%)~2D Graphics and Animation,
(26.7\%)~3D Graphics and Animation, and
(20.1\%)~User Interface,
(19.8\%)~Gameplay Logic.

\textbf{Complex and context-rich solutions.}
Similar to software tasks, \texttt{GameDevBench} solutions require multi-location edits that weave together multiple files. 
Our reference solutions average 4.7 files and 114.1 lines of code changed across 3.2 distinct filetypes (Figure~\ref{fig:statistics}, right).
This is more than triple the number of lines of code and file changes required compared to SWE-Bench~\citep{jimenez2024swebenchlanguagemodelsresolve}, suggesting substantial complexity to the tasks and corresponding solutions.

\textbf{Deterministic verification of multimodal solutions.}
Evaluating multimodal solutions is inherently challenging and solutions are typically evaluated through metrics such as CLIP~\citep{radford2021learningtransferablevisualmodels} or a Visual LLM-as-a-Judge~\citep{yin2026visionasinversegraphicsagentinterleavedmultimodal}.
These methods are, however, either proxies to correctness or non-deterministic.
\texttt{GameDevBench} instead uses Godot's testing framework which allows us to directly test game behavior.
For example, we can check to see if objects are in view of the camera or if object colliders have interacted purely through unit tests.
Thus, tests are repeatable and verifiable similar to software benchmarks while testing multimodal problems.

\textbf{Flexible solution methods.}
While the tests are deterministic, the methods used to derive solutions are flexible.
In this work, for simplicity, we evaluate agents that attempt to solve tasks through code generation alone.
However, it would be equally feasible to solve each task directly in the editor with approaches more similar to computer use.
Our test-based verification allows for direct comparison of different solution strategies.

\textbf{Continually Renewable.}
While not unique to our benchmark, our pipeline is repeatable and thus the benchmark can be continuously renewed. 
Human validation is minor with each task taking under 10 minutes to validate.

\section{Evaluation}

We evaluate various models and agentic harnesses on \texttt{GameDevBench}.

\textbf{Model Choices.}
From the Claude family of models we evaluate \model{Claude Haiku 4.5} and \model{Claude Sonnet 4.5}.
From the Gemini family we evaluate \model{Gemini 3 Flash} and \model{Gemini 3 Pro}.
From the GPT family we evaluate \model{GPT-5.4 Mini} and \model{GPT-5.4}.
For open weights models we evaluate \model{Qwen3.5-397B} and \model{Kimi K2.5}.

\textbf{Agent Harness Choices.}
To allow agents access to both the project files and the Godot application itself, we focus on agentic harnesses that operate locally.
We chose command-line interface (CLI) harnesses due to their ability to directly read code, image, and other asset files.
We evaluate each model in its respective agentic harness---\framework{claude-code} for Claude models, \framework{gemini-cli} for Gemini models, and \framework{codex} for ChatGPT models.
We evaluate \model{Kimi K2.5} using \framework{OpenHands}~\citep{wang2025openhandsopenplatformai}.
We also evaluate \model{Claude Haiku 4.5}, \model{Gemini 3 Flash}, and \model{GPT-5.4 Mini} using \framework{OpenHands} to compare performance across harnesses.

\subsection{Multimodal Feedback}

We describe two tooling configurations that allow agents to access richer multimodal information from Godot through editor screenshots and/or rendered video.


\textbf{Baseline.}
As a baseline, each agent starts inside the project directory and is given the task instruction along with basic instructions on how to run Godot.
We provide additional methods to support the agent, primarily to observe if additional visual context improves performance.
We provide our full prompts in Appendix~\ref{sec:prompt_templates}.

\textbf{Editor Screenshot MCP.}
We develop an MCP server that loads the Godot editor for the current task, takes a screenshot of the editor, then returns the image to the agent.
This allows the agent to view the game scene, the node tree, the node inspector, as well as other information present in the editor.
This method allows the agent to leverage additional visual feedback to validate its solution.

\textbf{Runtime Video.}
We provide agents with instructions on how to generate gameplay videos using Godot’s built-in recording functionality, which is otherwise frequently ignored or misused.
This differs from the MCP server as it captures both a) temporal elements only present in video and b) the current camera view (the editor does not show the camera view).
Typically, models process videos into image frames using python rather than ingesting the video directly.


\subsection{Discussion of Results}

\begin{figure}[h!]
    \centering
    \includegraphics[width=0.9\linewidth]{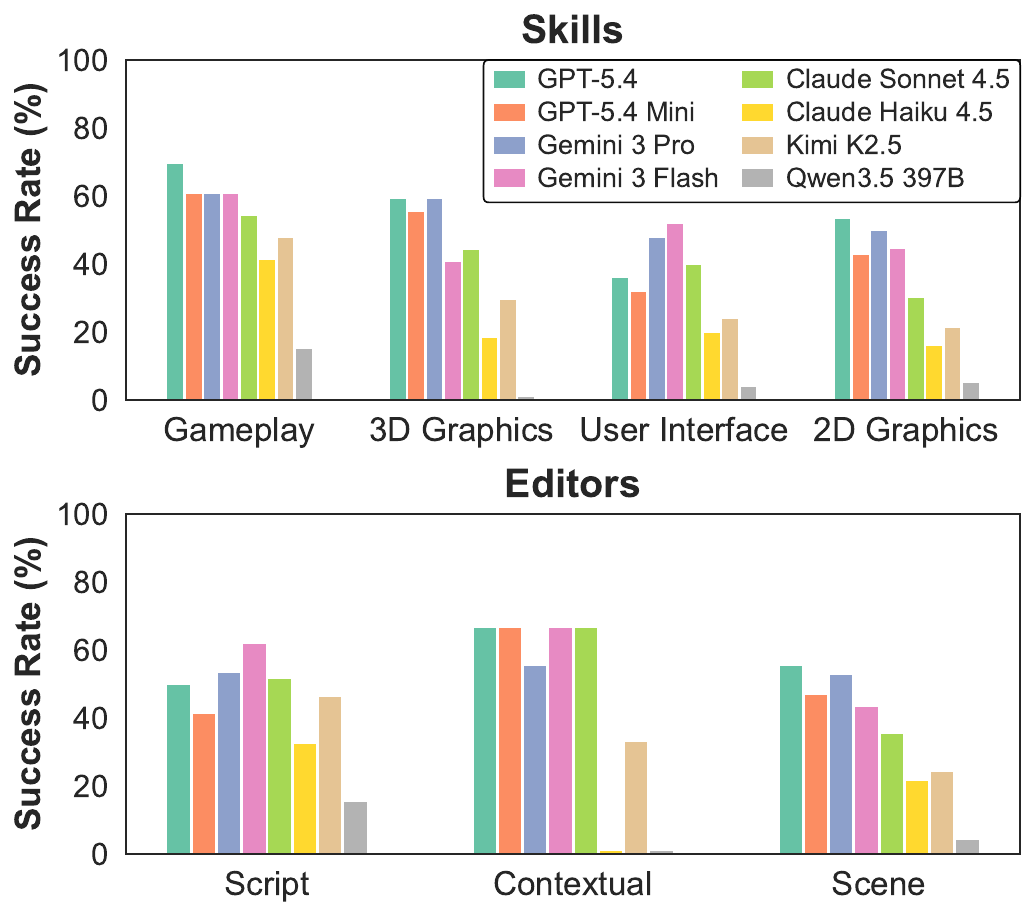}
    \caption{
    In general, agents perform better on tasks that require skills focusing on gameplay functionality compared to tasks that require multimodal understanding such as 2D and 3D graphics tasks.
    Performance across editor categories is dependent on the model: stronger frontier agents (\model{GPT-5.4 Mini}, \model{GPT-5.4}, \model{Gemini 3 Pro/Flash}, \model{Claude Sonnet 4.5}) maintain similar success across all editors, while weaker agents (\model{Claude Haiku 4.5}, \model{Kimi K2.5}, \model{Qwen3.5-397B}) drop sharply on the more multimodally demanding contextual and scene editors.
    All success rates are taken from results where the agent has access to multimodal feedback (MCP and Video).
    }
    \label{fig:domain}
\end{figure}

\begin{table*}[t!]
  \centering
  \begin{minipage}{0.95\textwidth}
  \centering
  \caption{Results from evaluating various models and agent harnesses on \texttt{GameDevBench}.
  Screenshot indicates that the agent was given access to an MCP server that screenshots the editor state.
  Video indicates that the agent was given additional instructions on how to generate a video of the current game scene.
  \textbf{Bold} and \textit{italics} indicate the best and second best model performance.
  }
  \label{tab:task_results}
  \small
  \begin{tabular}{l l c c c}
  \toprule
  Harness & Model & w/ Screenshot & w/ Video & \texttt{pass@1} (\%) \\
  \midrule
   \multirow{6}{*}{\centering \framework{claude-code}} & \multirow{4}{*}{\model{claude-haiku-4-5-20251001}} & \textcolor{red}{\texttimes} &
  \textcolor{red}{\texttimes} & $13.8$ \\
    &  & \textcolor{green}{\checkmark} & \textcolor{red}{\texttimes} & $15.6$ \\
    &  & \textcolor{red}{\texttimes} & \textcolor{green}{\checkmark} & $18.6$ \\
    &  & \textcolor{green}{\checkmark} & \textcolor{green}{\checkmark} & $16.5$ \\
  \cmidrule{2-5}
    & \multirow{2}{*}{\model{claude-sonnet-4-5-20250929}} & \textcolor{red}{\texttimes} & \textcolor{red}{\texttimes} & $28.8$ \\
    &  & \textcolor{green}{\checkmark} & \textcolor{green}{\checkmark} & $34.8$ \\
  \midrule
   \multirow{6}{*}{\centering \framework{codex}} & \multirow{4}{*}{\model{gpt-5.4-mini}} & \textcolor{red}{\texttimes} &
  \textcolor{red}{\texttimes} & $36.9$ \\
    &  & \textcolor{green}{\checkmark} & \textcolor{red}{\texttimes} & $37.8$ \\
    &  & \textcolor{red}{\texttimes} & \textcolor{green}{\checkmark} & $43.2$ \\
    &  & \textcolor{green}{\checkmark} & \textcolor{green}{\checkmark} & $39.0$ \\
  \cmidrule{2-5}
    & \multirow{2}{*}{\model{gpt-5.4}} & \textcolor{red}{\texttimes} & \textcolor{red}{\texttimes} & $41.1$ \\
    &  & \textcolor{green}{\checkmark} & \textcolor{green}{\checkmark} & $\mathit{52.0}$ \\
  \midrule
   \multirow{6}{*}{\centering \framework{gemini-cli}} & \multirow{4}{*}{\model{gemini-3-flash-preview}} & \textcolor{red}{\texttimes} &
  \textcolor{red}{\texttimes} & $45.4$ \\
    &  & \textcolor{green}{\checkmark} & \textcolor{red}{\texttimes} & $45.4$ \\
    &  & \textcolor{red}{\texttimes} & \textcolor{green}{\checkmark} & $46.9$ \\
    &  & \textcolor{green}{\checkmark} & \textcolor{green}{\checkmark} & $44.1$ \\
  \cmidrule{2-5}
    & \multirow{2}{*}{\model{gemini-3-pro-preview}} & \textcolor{red}{\texttimes} & \textcolor{red}{\texttimes} & $50.1$ \\
    &  & \textcolor{green}{\checkmark} & \textcolor{green}{\checkmark} & $\textbf{53.8}$ \\
  \midrule
   \multirow{10}{*}{\centering \framework{openhands}} & \multirow{2}{*}{\model{claude-haiku-4-5-20251001}} & \textcolor{red}{\texttimes} &
  \textcolor{red}{\texttimes} & $15.6$ \\
    &  & \textcolor{green}{\checkmark} & \textcolor{green}{\checkmark} & $17.7$ \\
  \cmidrule{2-5}
    & \multirow{2}{*}{\model{gpt-5.4-mini}} & \textcolor{red}{\texttimes} & \textcolor{red}{\texttimes} & $38.4$ \\
    &  & \textcolor{green}{\checkmark} & \textcolor{green}{\checkmark} & $36.9$ \\
  \cmidrule{2-5}
    & \multirow{2}{*}{\model{gemini-3-flash-preview}} & \textcolor{red}{\texttimes} & \textcolor{red}{\texttimes} & $30.3$ \\
    &  & \textcolor{green}{\checkmark} & \textcolor{green}{\checkmark} & $31.8$ \\
  \cmidrule{2-5}
    & \multirow{2}{*}{\model{kimi-k2.5}} & \textcolor{red}{\texttimes} & \textcolor{red}{\texttimes} & $18.9$ \\
    &  & \textcolor{green}{\checkmark} & \textcolor{green}{\checkmark} & $20.7$ \\
  \cmidrule{2-5}
    & \multirow{2}{*}{\model{qwen3.5-397b-a17b}} & \textcolor{red}{\texttimes} & \textcolor{red}{\texttimes} & $5.4$ \\
    &  & \textcolor{green}{\checkmark} & \textcolor{green}{\checkmark} & $5.1$ \\
  \bottomrule
  \end{tabular}
  \end{minipage}
\end{table*}

We now discuss our findings from evaluating agents on \texttt{GameDevBench} (Table~\ref{tab:task_results}).


\textbf{Game development proves challenging to even the most capable models and performance rapidly degrades when moving further from the frontier.} 
\model{GPT-5.4}, \model{Claude Sonnet 4.5}, and \model{Gemini 3 Pro} achieve baseline performances of $41.1\%$, $28.8\%$, and $50.1\%$ respectively without additional multimodal feedback in their native agentic harness.
Performance significantly degrades as we move further from the frontier.
Without multimodal support, \model{Claude Haiku 4.5} solves $13.8\%$, \model{Kimi K2.5} solves $18.9\%$, and \model{Qwen3.5-397B} solves only $5.4\%$ of tasks.
In contrast, \model{Kimi K2.5} solves $91.3\%$ of tasks in the frontend benchmark Design2Code~\citep{Si2024Design2CodeHFA, benchlm_design2code_2026}, illustrating that strong open-weights multimodal performance on web UI tasks does not transfer to game development.
Models struggle further on our \texttt{hard} subset of tasks which we discuss in more detail in Appendix~\ref{appendix:details}.



\textbf{Agent performance differs significantly across skill and editor categories.}
We observe a general trend where agents perform worse on tasks that are more multimodally demanding (Figure~\ref{fig:domain}).
For skills, agents perform best at gameplay logic tasks ($51.4\%$) and worst at 2D graphics and animation ($33.0\%$) and UI ($32.0\%$) tasks, which require agents to understand images or other assets for animations and effects. 
3D graphics ($38.4\%$) sit in between (scores are averages computed across the 8 agents in their native agentic harness with multimodal feedback enabled).
Performance across editor categories is instead dependent on model capabilities.
The models that achieve over 30\% \texttt{pass@1}---\model{GPT-5.4 Mini}, \model{GPT-5.4}, \model{Gemini 3 Pro}, \model{Gemini 3 Flash}, and \model{Claude Sonnet 4.5}---perform similarly on tasks regardless of the required editor type.
However, \model{Claude Haiku 4.5}, \model{Kimi K2.5}, and \model{Qwen3.5-397B} perform worse on tasks requiring the scene and contextual editors which are typically more multimodally demanding compared to scripting tasks.

\begin{figure*}[h!]
    \centering
    \includegraphics[width=0.85\linewidth]{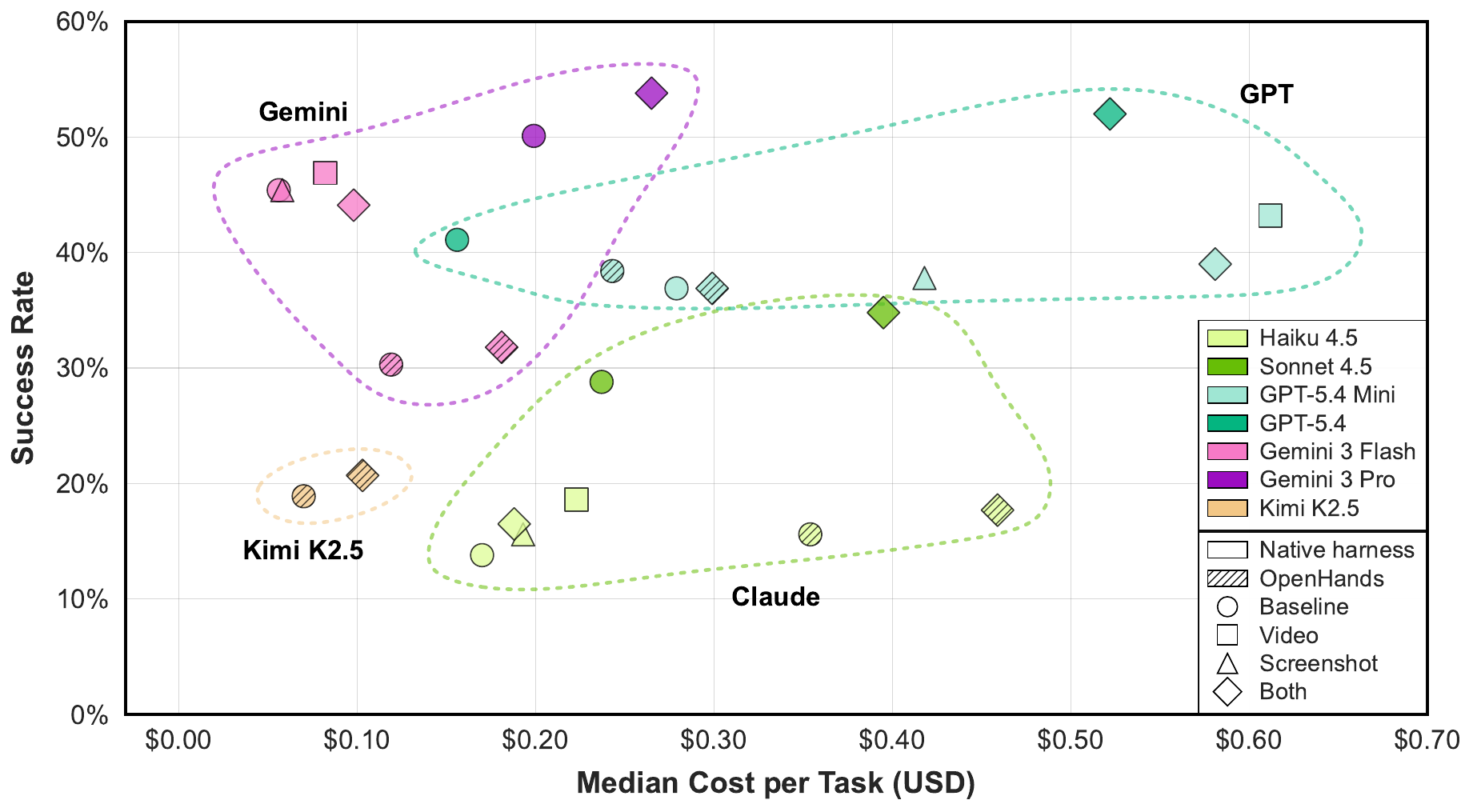}
    \caption{We capture the trade-off between performance and cost.
    We observe that Gemini models are the most cost efficient, offering good performance at lower costs.
    Among all the models, GPT costs vary the most when utilizing multimodal feedback.
    Comparatively, Claude models generally under-perform in both cost and performance.
    }
    \label{fig:cost}
\end{figure*}

\textbf{Multimodal tooling consistently improves agent performance.}
We find that providing an agent with both the MCP and video instructions almost always improves performance.
This trend holds across most models, though the gains vary across agents.
We see the largest gain with \model{GPT-5.4} where performance increases from $41.1\% \rightarrow 52.0\%$.
The only exceptions to this are \model{GPT-5.4 Mini} in \framework{OpenHands}---which is not its native harness---and \model{Gemini 3 Flash} where performance marginally decreases. 
Additionally, we evaluate \model{GPT-5.4 Mini}, \model{Claude Haiku 4.5}, and \model{Gemini 3 Flash} using the MCP or video instructions in isolation.
Surprisingly, video runtime often provides more improvement than enabling both forms of multimodal feedback.
An exception is \model{Gemini 3 Flash} which performs similarly across all methods.
Although all tasks can be verified through code, visual feedback allows agents to verify and amend mistakes.
This behavior strongly resembles that seen in recent work~\citep{yin2026visionasinversegraphicsagentinterleavedmultimodal}, where visual feedback improves agentic performance.

\textbf{Agentic harness choice can impact performance, but the effect varies depending on the model.}
We evaluate \model{Claude Haiku 4.5}, \model{Gemini 3 Flash}, and \model{GPT-5.4 Mini} using both their original harness and \framework{OpenHands} (Table~\ref{tab:task_results}).
When using \framework{OpenHands}, we observe that \model{Claude Haiku 4.5} shows a small increase in performance from $16.5\%$ to $17.7\%$.
\model{GPT-5.4 Mini} offers mixed performance, improving over the baseline from $36.9\%$ to $38.4\%$, yet degrading from $39.0\%$ to $36.9\%$ when using multimodal feedback in \framework{OpenHands}. 
On the other hand, \model{Gemini 3 Flash}'s performance significantly decreases from $45.4\%$ to $30.3\%$ in \framework{OpenHands}.
This is likely due to incompatible editing tools between Gemini models and \framework{OpenHands}~\footnote{https://github.com/OpenHands/OpenHands/issues/9454}.

\textbf{Cost varies significantly depending on the model, harness, and whether multimodal feedback is provided.}
We find that enabling multimodal feedback almost always increases cost in exchange for increased performance (Figure~\ref{fig:cost}).
When enabling multimodal feedback, both \model{Gemini} and \model{Claude} models tend to show a relatively minor increase in costs.
On the other hand, \model{GPT} models increase significantly in cost, with up to a 3.3x increase in \model{GPT-5.4} when enabling feedback.
Interestingly, costs when using multimodal feedback does not increase significantly when \model{GPT-5.4 Mini} is used in \framework{OpenHands}, although neither does performance.
We find that \model{Gemini 3 Flash} in its native harness is the most cost-efficient model.

\subsection{Error Analysis and Directions for Improvement.}

We manually analyzed some of the most common errors that agents made when solving the task.
These errors indicate potential gaps in capabilities and future directions for agent development.
While there are a variety of errors, we observe two consistent error patterns.
We provide a more detailed analysis on error patterns in Appendix~\ref{app:failure_analysis}.

\textbf{Agents struggle with multimodal understanding.}
Perhaps the most consistent error pattern occurs from a lack of multimodal understanding.
Specifically, it is often necessary to understand multimodal inputs to properly complete a game development task.
For example, creating (or even simply picking) an animation requires that the agent either parse through multiple images or pick out specific sprites within a spritesheet.
Currently, agents frequently pick the wrong images or sprites (e.g., picking walking motion sprites instead of attacking motions).
It is clear that improvements to multimodal understanding would significantly improve performance of agentic game development.

\textbf{Agents struggle with common game development patterns.} 
In game development, there are many common development patterns.
For example, game elements (called nodes in Godot) form a tree structure where specific nodes such as an AnimatedSprite2D and CapsuleCollider handle animations and physics respectively.
Another example would be signals that trigger between various files when conditions are met such as when two colliders intersect with each other.
Agents frequently add nodes to incorrect levels in the tree, drop necessary signals, or assign resources to the wrong elements.
We provide an example of such an error in Appendix~\ref{appendix:case_study}.
This reinforces a long-standing trend within model training---models must be trained on specific domains to excel within that domain.

\section{Related Works}


\textbf{Agentic Benchmarks.}
Software development has been one of the premier frontiers of agentic development.
SWE-Bench~\citep{jimenez2024swebenchlanguagemodelsresolve, yang2024sweagent} was perhaps the first benchmark and catalyst towards agentic software development.
Over time, multiple new software benchmarks have been developed~\citep{chan2025mlebenchevaluatingmachinelearning, merrill2026terminalbenchbenchmarkingagentshard, yang2025codeclashbenchmarkinggoalorientedsoftware}, but they remain largely unimodal.
The few multimodal software benchmarks have largely focused on frontend JavaScript development~\citep{Zhu2025FrontendBenchABA,Si2024Design2CodeHFA,yang2024swebenchmultimodalaisystems}.
Instead, the most common use case for multimodal agents has been computer use~\citep{xie2024osworldbenchmarkingmultimodalagents} and web navigation~\citep{zhou2024webarenarealisticwebenvironment, koh2024visualwebarenaevaluatingmultimodalagents}. 
Progress in this domain is challenging as agents must operate in an action space rather than simply writing code.
Game development bridges the gaps between these domains by requiring multimodal input, but allowing for code output.
\texttt{GameDevBench} is able to effectively reap benefits from both software and computer use domains, thus enabling effective multimodal evaluation.

\textbf{Game Playing.} 
There has always been significant interest in the application of artificial intelligence (AI) to games~\citep{gallotta2024large};
gameplay has been seen as a proxy for the capabilities or intelligence of an AI system, with projects ranging from Deep Blue~\citep{campbell2002deep}, Alpha Go~\citep{silver2016mastering}, and Cicero~\citep{meta2022human} to more recent generalists such as SIMA 2~\citep{bolton2025sima}. 
Practically, games provide an interactive simulation environment with clear reward signals allowing researchers to experiment with methods---particularly from reinforcement learning---to improve model capabilities.
The recent flux of LLMs playing Pokémon~\citep{karten2025pokeagent,karten2025pok,comanici2025gemini} uses game-playing agents to evaluate and explore the agentic reasoning capabilities of frontier models which is then used directly in game development to test games~\citep{nunuai2024pokemon}.
This transition from game playing agents as NPCs and opponents in games to becoming a portion of the game development process marks a timely need for benchmarks such as \texttt{GameDevBench}.


\textbf{Game Development.}
Concordia~\citep{vezhnevets2023generative} and other subsequent work on tabletop role-playing games~\citep{vezhnevets2025multi} seek to replace interactable characters with a highly adaptive story created entirely from interactions with LLMs.
Other works try to fully replace the physics engine of the game to immediately generate frames based on player actions~\citep{bruce2024genie}.
Procedural Content Generation has a long history of using AI for game asset creation~\citep{summerville2018proceduralcontentgenerationmachine, shaker2016procedural} and evolutionary level design~\citep{sudhakaran2023mariogptopenendedtext2levelgeneration}.
However, these largely focus on a singular aspect of game development.
Ultimately, each of these features still needs to be combined in a game engine to develop a full game, which is the capability \texttt{GameDevBench} directly evaluates.

    
    

\section{Conclusion}

We present \texttt{GameDevBench}, the first benchmark to evaluate an agent's ability to solve game development tasks.
To create our benchmark, we develop a pipeline for converting YouTube and web tutorials into benchmark tasks.
We find that agents struggle with tasks in game development, especially when tasks require deeper multimodal understanding.
The gap between frontier and non-frontier models is sharp, with absolute differences of up to $48.7\%$ \texttt{pass@1}.
Lastly, we show that even simple multimodal feedback tooling improves agent performance: when given access to screenshots and video, \model{GPT-5.4} increases from $41.1\%$ to $52.0\%$ \texttt{pass@1}, a $10.9$ percentage-point gain and a $26.5\%$ relative improvement.
Our findings highlight the need to improve multimodal capabilities of agents---either through training or methods of visual feedback.
We speculate addressing these needs would improve agentic performance in domains even beyond software and game development.

\newpage
\section*{Acknowledgements}

This work was supported in part by the National Science Foundation grants IIS1705121, IIS1838017, IIS2046613, IIS2112471, the Department of Defense (DoD) through the National Defense Science and Engineering Graduate (NDSEG) Fellowship Program, and funding from Datadog. Any opinions, findings and conclusions or recommendations expressed in this material are those of the author(s) and do not necessarily reflect the views of any of these funding agencies.
\section*{Impact Statement}

This work introduces \texttt{GameDevBench}, a benchmark for evaluating agentic capabilities in game development, a domain that combines large-scale software engineering with rich multimodal reasoning. 

\textbf{Positive impacts.} 
\texttt{GameDevBench} may accelerate the development of more capable and robust multimodal agents, with downstream benefits extending beyond game development to other visually grounded, context-dense domains such as robotics, simulation, design tools, and interactive software engineering. 
Additionally, by focusing on an open-source engine (Godot) and releasing tasks and infrastructure publicly, this work lowers barriers to entry and supports reproducible research by both academic and independent researchers.

\textbf{Potential risks and misuse.}
Game development is both a software discipline and a creative discipline.
Improved agentic capabilities could contribute to workforce disruption in creative and technical fields either directly in game development or beyond. 
While \texttt{GameDevBench} does not aim to replace human developers, advances driven by this benchmark may enable partial automation of tasks traditionally performed by artists or programmers. 

\textbf{Mitigations and ethical considerations.} 
To reduce potential risks, tasks are derived from permissively licensed tutorials, and no proprietary assets or data are included. 
Additionally, none of the tasks involve \textit{creating} assets themselves, instead focusing exclusively on tasks that involve the use of \textit{pre-existing} assets, thus limiting direct impact to artists and creatives.

\bibliography{example_paper}

@misc{zhou2024webarenarealisticwebenvironment,
      title={WebArena: A Realistic Web Environment for Building Autonomous Agents}, 
      author={Shuyan Zhou and Frank F. Xu and Hao Zhu and Xuhui Zhou and Robert Lo and Abishek Sridhar and Xianyi Cheng and Tianyue Ou and Yonatan Bisk and Daniel Fried and Uri Alon and Graham Neubig},
      year={2024},
      eprint={2307.13854},
      archivePrefix={arXiv},
      primaryClass={cs.AI},
      url={https://arxiv.org/abs/2307.13854}, 
}

@inproceedings{yang2024sweagent,
  title={{SWE}-agent: Agent-Computer Interfaces Enable Automated Software Engineering},
  author={John Yang and Carlos E Jimenez and Alexander Wettig and Kilian Lieret and Shunyu Yao and Karthik R Narasimhan and Ofir Press},
  booktitle={The Thirty-eighth Annual Conference on Neural Information Processing Systems},
  year={2024},
  url={https://arxiv.org/abs/2405.15793}
}

@misc{zheng2023judgingllmasajudgemtbenchchatbot,
      title={Judging LLM-as-a-Judge with MT-Bench and Chatbot Arena}, 
      author={Lianmin Zheng and Wei-Lin Chiang and Ying Sheng and Siyuan Zhuang and Zhanghao Wu and Yonghao Zhuang and Zi Lin and Zhuohan Li and Dacheng Li and Eric P. Xing and Hao Zhang and Joseph E. Gonzalez and Ion Stoica},
      year={2023},
      eprint={2306.05685},
      archivePrefix={arXiv},
      primaryClass={cs.CL},
      url={https://arxiv.org/abs/2306.05685}, 
}

@article{jagli2024artificial,
  title={Artificial Intelligence Usage in Game Development},
  author={Jagli, Dhanamma and Nalla, Subhashchandra and Danikonda, Srinivasrao and Nakirekanti, Laxmi},
  year={2024},
  publisher={Preprints}
}

@article{filipovic2023role,
  title={The role of artificial intelligence in video game development},
  author={Filipovi{\'c}, Aleksandar},
  journal={Kultura polisa},
  volume={20},
  number={3},
  pages={50--67},
  year={2023},
  publisher={Удружење „КУЛТУРА--ПОЛИС “Нови Сад}
}

@article{yakan2022analysis,
  title={Analysis of development of artificial intelligence in the game industry},
  author={Yakan, Sayid Adli},
  journal={International Journal of Cyber and IT Service Management},
  volume={2},
  number={2},
  pages={111--116},
  year={2022}
}

@misc{jimenez2024swebenchlanguagemodelsresolve,
      title={SWE-bench: Can Language Models Resolve Real-World GitHub Issues?}, 
      author={Carlos E. Jimenez and John Yang and Alexander Wettig and Shunyu Yao and Kexin Pei and Ofir Press and Karthik Narasimhan},
      year={2024},
      eprint={2310.06770},
      archivePrefix={arXiv},
      primaryClass={cs.CL},
      url={https://arxiv.org/abs/2310.06770}, 
}

@misc{chi2025editbenchevaluatingllmabilities,
      title={EDIT-Bench: Evaluating LLM Abilities to Perform Real-World Instructed Code Edits}, 
      author={Wayne Chi and Valerie Chen and Ryan Shar and Aditya Mittal and Jenny Liang and Wei-Lin Chiang and Anastasios Nikolas Angelopoulos and Ion Stoica and Graham Neubig and Ameet Talwalkar and Chris Donahue},
      year={2025},
      eprint={2511.04486},
      archivePrefix={arXiv},
      primaryClass={cs.SE},
      url={https://arxiv.org/abs/2511.04486}, 
}

@misc{koh2024visualwebarenaevaluatingmultimodalagents,
      title={VisualWebArena: Evaluating Multimodal Agents on Realistic Visual Web Tasks}, 
      author={Jing Yu Koh and Robert Lo and Lawrence Jang and Vikram Duvvur and Ming Chong Lim and Po-Yu Huang and Graham Neubig and Shuyan Zhou and Ruslan Salakhutdinov and Daniel Fried},
      year={2024},
      eprint={2401.13649},
      archivePrefix={arXiv},
      primaryClass={cs.LG},
      url={https://arxiv.org/abs/2401.13649}, 
}

@misc{xie2024osworldbenchmarkingmultimodalagents,
      title={OSWorld: Benchmarking Multimodal Agents for Open-Ended Tasks in Real Computer Environments}, 
      author={Tianbao Xie and Danyang Zhang and Jixuan Chen and Xiaochuan Li and Siheng Zhao and Ruisheng Cao and Toh Jing Hua and Zhoujun Cheng and Dongchan Shin and Fangyu Lei and Yitao Liu and Yiheng Xu and Shuyan Zhou and Silvio Savarese and Caiming Xiong and Victor Zhong and Tao Yu},
      year={2024},
      eprint={2404.07972},
      archivePrefix={arXiv},
      primaryClass={cs.AI},
      url={https://arxiv.org/abs/2404.07972}, 
}

@misc{yin2026visionasinversegraphicsagentinterleavedmultimodal,
      title={Vision-as-Inverse-Graphics Agent via Interleaved Multimodal Reasoning}, 
      author={Shaofeng Yin and Jiaxin Ge and Zora Zhiruo Wang and Xiuyu Li and Michael J. Black and Trevor Darrell and Angjoo Kanazawa and Haiwen Feng},
      year={2026},
      eprint={2601.11109},
      archivePrefix={arXiv},
      primaryClass={cs.CV},
      url={https://arxiv.org/abs/2601.11109}, 
}

@article{Si2024Design2CodeHFA,
  title={Design2Code: How Far Are We From Automating Front-End Engineering?},
  author={Chenglei Si and Yanzhe Zhang and Ryan Li and Zhengyuan Yang and Ruibo Liu and Diyi Yang},
  journal={ArXiv},
  year={2024},
  volume={abs/2403.03163},
  url={https://api.semanticscholar.org/CorpusId:268248801}
}

@article{Zhu2025FrontendBenchABA,
  title={FrontendBench: A Benchmark for Evaluating LLMs on Front-End Development via Automatic Evaluation},
  author={Hongda Zhu and Yiwen Zhang and Bing Zhao and Jingzhe Ding and Siyao Liu and Tong Liu and Dandan Wang and Yanan Liu and Zhaojian Li},
  journal={ArXiv},
  year={2025},
  volume={abs/2506.13832},
  url={https://api.semanticscholar.org/CorpusId:279410903}
}

@misc{yang2024swebenchmultimodalaisystems,
      title={SWE-bench Multimodal: Do AI Systems Generalize to Visual Software Domains?}, 
      author={John Yang and Carlos E. Jimenez and Alex L. Zhang and Kilian Lieret and Joyce Yang and Xindi Wu and Ori Press and Niklas Muennighoff and Gabriel Synnaeve and Karthik R. Narasimhan and Diyi Yang and Sida I. Wang and Ofir Press},
      year={2024},
      eprint={2410.03859},
      archivePrefix={arXiv},
      primaryClass={cs.CL},
      url={https://arxiv.org/abs/2410.03859}, 
}

@article{Vinyals2019GrandmasterLIA,
  title={Grandmaster level in StarCraft II using multi-agent reinforcement learning},
  author={O. Vinyals and Igor Babuschkin and Wojciech M. Czarnecki and Micha{\"e}l Mathieu and A. Dudzik and Junyoung Chung and David Choi and Richard Powell and T. Ewalds and Petko Georgiev and Junhyuk Oh and Dan Horgan and M. Kroiss and Ivo Danihelka and Aja Huang and L. Sifre and Trevor Cai and J. Agapiou and Max Jaderberg and A. Vezhnevets and R{\'e}mi Leblond and Tobias Pohlen and Valentin Dalibard and D. Budden and Yury Sulsky and James Molloy and T. Paine and Caglar Gulcehre and Ziyun Wang and T. Pfaff and Yuhuai Wu and Roman Ring and Dani Yogatama and Dario W{\"u}nsch and Katrina McKinney and Oliver Smith and T. Schaul and T. Lillicrap and K. Kavukcuoglu and D. Hassabis and C. Apps and David Silver},
  journal={Nature},
  year={2019},
  volume={575},
  pages={350 - 354},
  url={https://doi.org/10.1038/s41586-019-1724-z}
}

@article{Schrittwieser2019MasteringAGA,
  title={Mastering Atari, Go, chess and shogi by planning with a learned model},
  author={Julian Schrittwieser and Ioannis Antonoglou and T. Hubert and K. Simonyan and L. Sifre and Simon Schmitt and A. Guez and Edward Lockhart and D. Hassabis and T. Graepel and T. Lillicrap and David Silver},
  journal={Nature},
  year={2019},
  volume={588},
  pages={604 - 609},
  url={https://doi.org/10.1038/s41586-020-03051-4}
}

@article{silver2016mastering,
  title={Mastering the game of Go with deep neural networks and tree search},
  author={Silver, David and Huang, Aja and Maddison, Chris J and Guez, Arthur and Sifre, Laurent and Van Den Driessche, George and Schrittwieser, Julian and Antonoglou, Ioannis and Panneershelvam, Veda and Lanctot, Marc and others},
  journal={nature},
  volume={529},
  number={7587},
  pages={484--489},
  year={2016},
  publisher={Nature Publishing Group}
}

@article{Silver2018AGRA,
  title={A general reinforcement learning algorithm that masters chess, shogi, and Go through self-play},
  author={David Silver and T. Hubert and Julian Schrittwieser and Ioannis Antonoglou and Matthew Lai and A. Guez and Marc Lanctot and L. Sifre and D. Kumaran and T. Graepel and T. Lillicrap and K. Simonyan and D. Hassabis},
  journal={Science},
  year={2018},
  volume={362},
  pages={1140 - 1144},
  url={https://doi.org/10.1126/science.aar6404}
}

@article{Valevski2024DiffusionMAA,
  title={Diffusion Models Are Real-Time Game Engines},
  author={Dani Valevski and Yaniv Leviathan and Moab Arar and Shlomi Fruchter},
  journal={ArXiv},
  year={2024},
  volume={abs/2408.14837},
  url={https://api.semanticscholar.org/CorpusId:271962839}
}

@inproceedings{Oh2015ActionConditionalVPA,
  title={Action-Conditional Video Prediction using Deep Networks in Atari Games},
  author={Junhyuk Oh and Xiaoxiao Guo and Honglak Lee and Richard L. Lewis and Satinder Singh},
  booktitle={Neural Information Processing Systems},
  year={2015},
  url={https://api.semanticscholar.org/CorpusId:3147510}
}

@misc{anthropic2024modelcontext,
  title        = {Introducing the Model Context Protocol},
  author       = {{Anthropic}},
  year         = {2024},
  month        = nov,
  day          = {25},
  howpublished = {\url{https://www.anthropic.com/news/model-context-protocol}}
}

@article{karten2025pokeagent,
  title={The pokeagent challenge: Competitive and long-context learning at scale},
  author={Karten, Seth and Grigsby, Jake and Milani, Stephanie and Vodrahalli, Kiran and Zhang, Amy and Fang, Fei and Zhu, Yuke and Jin, Chi},
  journal={NeurIPS Competition Track},
  year={2025}
}

@article{karten2025pok,
  title={PokéChamp: an Expert-level Minimax Language Agent},
  author={Karten, Seth and Nguyen, Andy Luu and Jin, Chi},
  journal={arXiv preprint arXiv:2503.04094},
  year={2025}
}

@article{comanici2025gemini,
  title={Gemini 2.5: Pushing the frontier with advanced reasoning, multimodality, long context, and next generation agentic capabilities},
  author={Comanici, Gheorghe and Bieber, Eric and Schaekermann, Mike and Pasupat, Ice and Sachdeva, Noveen and Dhillon, Inderjit and Blistein, Marcel and Ram, Ori and Zhang, Dan and Rosen, Evan and others},
  journal={arXiv preprint arXiv:2507.06261},
  year={2025}
}

@article{vezhnevets2025multi,
  title={Multi-Actor Generative Artificial Intelligence as a Game Engine},
  author={Vezhnevets, Alexander Sasha and Matyas, Jayd and Cross, Logan and Paglieri, Davide and Chang, Minsuk and Cunningham, William A and Osindero, Simon and Isaac, William S and Leibo, Joel Z},
  journal={arXiv preprint arXiv:2507.08892},
  year={2025}
}

@article{vezhnevets2023generative,
  title={Generative agent-based modeling with actions grounded in physical, social, or digital space using Concordia},
  author={Vezhnevets, Alexander Sasha and Agapiou, John P and Aharon, Avia and Ziv, Ron and Matyas, Jayd and Du{\'e}{\~n}ez-Guzm{\'a}n, Edgar A and Cunningham, William A and Osindero, Simon and Karmon, Danny and Leibo, Joel Z},
  journal={arXiv preprint arXiv:2312.03664},
  year={2023}
}

@article{meta2022human,
  title={Human-level play in the game of Diplomacy by combining language models with strategic reasoning},
  author={FAIR and Bakhtin, Anton and Brown, Noam and Dinan, Emily and Farina, Gabriele and Flaherty, Colin and Fried, Daniel and Goff, Andrew and Gray, Jonathan and Hu, Hengyuan and others},
  journal={Science},
  volume={378},
  number={6624},
  pages={1067--1074},
  year={2022},
  publisher={American Association for the Advancement of Science}
}

@article{bolton2025sima,
  title={Sima 2: A generalist embodied agent for virtual worlds},
  author={Bolton, Adrian and Lerchner, Alexander and Cordell, Alexandra and Moufarek, Alexandre and Bolt, Andrew and Lampinen, Andrew and Mitenkova, Anna and Hallingstad, Arne Olav and Vujatovic, Bojan and Li, Bonnie and others},
  journal={arXiv preprint arXiv:2512.04797},
  year={2025}
}

@inproceedings{bruce2024genie,
  title={Genie: Generative interactive environments},
  author={Bruce, Jake and Dennis, Michael D and Edwards, Ashley and Parker-Holder, Jack and Shi, Yuge and Hughes, Edward and Lai, Matthew and Mavalankar, Aditi and Steigerwald, Richie and Apps, Chris and others},
  booktitle={Forty-first International Conference on Machine Learning},
  year={2024}
}

@misc{nunuai2024pokemon,
  author       = {{Nunu AI}},
  title        = {Beating the World Record in Pokémon Emerald: An {AI} Agent Case Study},
  howpublished = {\url{https://nunu.ai/case-studies/pokemon-emerald}},
  year         = {2024}
}

@article{gallotta2024large,
  title={Large language models and games: A survey and roadmap},
  author={Gallotta, Roberto and Todd, Graham and Zammit, Marvin and Earle, Sam and Liapis, Antonios and Togelius, Julian and Yannakakis, Georgios N},
  journal={IEEE Transactions on Games},
  year={2024},
  publisher={IEEE}
}

@misc{wang2025openhandsopenplatformai,
      title={OpenHands: An Open Platform for AI Software Developers as Generalist Agents}, 
      author={Xingyao Wang and Boxuan Li and Yufan Song and Frank F. Xu and Xiangru Tang and Mingchen Zhuge and Jiayi Pan and Yueqi Song and Bowen Li and Jaskirat Singh and Hoang H. Tran and Fuqiang Li and Ren Ma and Mingzhang Zheng and Bill Qian and Yanjun Shao and Niklas Muennighoff and Yizhe Zhang and Binyuan Hui and Junyang Lin and Robert Brennan and Hao Peng and Heng Ji and Graham Neubig},
      year={2025},
      eprint={2407.16741},
      archivePrefix={arXiv},
      primaryClass={cs.SE},
      url={https://arxiv.org/abs/2407.16741}, 
}

@misc{kidscancode_godot_recipes_4x,
  author       = {{KidsCanCode}},
  title        = {{Godot Recipes}},
  howpublished = {\url{https://kidscancode.org/godot_recipes/4.x/}},
  note         = {Version 4.x, accessed January 28, 2026}
}

@misc{chan2025mlebenchevaluatingmachinelearning,
      title={MLE-bench: Evaluating Machine Learning Agents on Machine Learning Engineering}, 
      author={Jun Shern Chan and Neil Chowdhury and Oliver Jaffe and James Aung and Dane Sherburn and Evan Mays and Giulio Starace and Kevin Liu and Leon Maksin and Tejal Patwardhan and Lilian Weng and Aleksander Mądry},
      year={2025},
      eprint={2410.07095},
      archivePrefix={arXiv},
      primaryClass={cs.CL},
      url={https://arxiv.org/abs/2410.07095}, 
}

@misc{merrill2026terminalbenchbenchmarkingagentshard,
      title={Terminal-Bench: Benchmarking Agents on Hard, Realistic Tasks in Command Line Interfaces}, 
      author={Mike A. Merrill and Alexander G. Shaw and Nicholas Carlini and Boxuan Li and Harsh Raj and Ivan Bercovich and Lin Shi and Jeong Yeon Shin and Thomas Walshe and E. Kelly Buchanan and Junhong Shen and Guanghao Ye and Haowei Lin and Jason Poulos and Maoyu Wang and Marianna Nezhurina and Jenia Jitsev and Di Lu and Orfeas Menis Mastromichalakis and Zhiwei Xu and Zizhao Chen and Yue Liu and Robert Zhang and Leon Liangyu Chen and Anurag Kashyap and Jan-Lucas Uslu and Jeffrey Li and Jianbo Wu and Minghao Yan and Song Bian and Vedang Sharma and Ke Sun and Steven Dillmann and Akshay Anand and Andrew Lanpouthakoun and Bardia Koopah and Changran Hu and Etash Guha and Gabriel H. S. Dreiman and Jiacheng Zhu and Karl Krauth and Li Zhong and Niklas Muennighoff and Robert Amanfu and Shangyin Tan and Shreyas Pimpalgaonkar and Tushar Aggarwal and Xiangning Lin and Xin Lan and Xuandong Zhao and Yiqing Liang and Yuanli Wang and Zilong Wang and Changzhi Zhou and David Heineman and Hange Liu and Harsh Trivedi and John Yang and Junhong Lin and Manish Shetty and Michael Yang and Nabil Omi and Negin Raoof and Shanda Li and Terry Yue Zhuo and Wuwei Lin and Yiwei Dai and Yuxin Wang and Wenhao Chai and Shang Zhou and Dariush Wahdany and Ziyu She and Jiaming Hu and Zhikang Dong and Yuxuan Zhu and Sasha Cui and Ahson Saiyed and Arinbjörn Kolbeinsson and Jesse Hu and Christopher Michael Rytting and Ryan Marten and Yixin Wang and Alex Dimakis and Andy Konwinski and Ludwig Schmidt},
      year={2026},
      eprint={2601.11868},
      archivePrefix={arXiv},
      primaryClass={cs.SE},
      url={https://arxiv.org/abs/2601.11868}, 
}

@misc{yang2025codeclashbenchmarkinggoalorientedsoftware,
      title={CodeClash: Benchmarking Goal-Oriented Software Engineering}, 
      author={John Yang and Kilian Lieret and Joyce Yang and Carlos E. Jimenez and Ofir Press and Ludwig Schmidt and Diyi Yang},
      year={2025},
      eprint={2511.00839},
      archivePrefix={arXiv},
      primaryClass={cs.SE},
      url={https://arxiv.org/abs/2511.00839}, 
}

@misc{sudhakaran2023mariogptopenendedtext2levelgeneration,
      title={MarioGPT: Open-Ended Text2Level Generation through Large Language Models}, 
      author={Shyam Sudhakaran and Miguel González-Duque and Claire Glanois and Matthias Freiberger and Elias Najarro and Sebastian Risi},
      year={2023},
      eprint={2302.05981},
      archivePrefix={arXiv},
      primaryClass={cs.AI},
      url={https://arxiv.org/abs/2302.05981}, 
}

@misc{summerville2018proceduralcontentgenerationmachine,
      title={Procedural Content Generation via Machine Learning (PCGML)}, 
      author={Adam Summerville and Sam Snodgrass and Matthew Guzdial and Christoffer Holmgård and Amy K. Hoover and Aaron Isaksen and Andy Nealen and Julian Togelius},
      year={2018},
      eprint={1702.00539},
      archivePrefix={arXiv},
      primaryClass={cs.AI},
      url={https://arxiv.org/abs/1702.00539}, 
}

@article{shaker2016procedural,
  title={Procedural content generation in games},
  author={Shaker, Noor and Togelius, Julian and Nelson, Mark J},
  year={2016},
  publisher={Springer}
}

@article{campbell2002deep,
  title={Deep blue},
  author={Campbell, Murray and Hoane Jr, A Joseph and Hsu, Feng-hsiung},
  journal={Artificial intelligence},
  volume={134},
  number={1-2},
  pages={57--83},
  year={2002},
  publisher={Elsevier}
}

@misc{radford2021learningtransferablevisualmodels,
      title={Learning Transferable Visual Models From Natural Language Supervision}, 
      author={Alec Radford and Jong Wook Kim and Chris Hallacy and Aditya Ramesh and Gabriel Goh and Sandhini Agarwal and Girish Sastry and Amanda Askell and Pamela Mishkin and Jack Clark and Gretchen Krueger and Ilya Sutskever},
      year={2021},
      eprint={2103.00020},
      archivePrefix={arXiv},
      primaryClass={cs.CV},
      url={https://arxiv.org/abs/2103.00020}, 
}

@misc{wang2023largelanguagemodelsfair,
      title={Large Language Models are not Fair Evaluators}, 
      author={Peiyi Wang and Lei Li and Liang Chen and Zefan Cai and Dawei Zhu and Binghuai Lin and Yunbo Cao and Qi Liu and Tianyu Liu and Zhifang Sui},
      year={2023},
      eprint={2305.17926},
      archivePrefix={arXiv},
      primaryClass={cs.CL},
      url={https://arxiv.org/abs/2305.17926}, 
}

@misc{koo2024benchmarkingcognitivebiaseslarge,
      title={Benchmarking Cognitive Biases in Large Language Models as Evaluators}, 
      author={Ryan Koo and Minhwa Lee and Vipul Raheja and Jong Inn Park and Zae Myung Kim and Dongyeop Kang},
      year={2024},
      eprint={2309.17012},
      archivePrefix={arXiv},
      primaryClass={cs.CL},
      url={https://arxiv.org/abs/2309.17012}, 
}

@misc{benchlm_design2code_2026,
  title        = {{Design2Code}},
  author       = {{BenchLM.ai}},
  year         = {2026},
  howpublished = {\url{https://benchlm.ai/benchmarks/design2Code}},
  note         = {Benchmark page. Last updated May 27, 2026. Accessed May 28, 2026}
}

@misc{liang2025slidegencollaborativemultimodalagents,
      title={SlideGen: Collaborative Multimodal Agents for Scientific Slide Generation}, 
      author={Xin Liang and Xiang Zhang and Yiwei Xu and Siqi Sun and Chenyu You},
      year={2025},
      eprint={2512.04529},
      archivePrefix={arXiv},
      primaryClass={cs.AI},
      url={https://arxiv.org/abs/2512.04529}, 
}
\bibliographystyle{icml2026}


\newpage
\appendix
\onecolumn
\appendix

\section{Task Construction Prompt}\label{appendix:task_construction}

Below is the full prompt provided to the Codex agent for automatic task construction from YouTube tutorials (Stage 2). The agent receives this prompt along with a pointer to a specific tutorial folder containing a video transcript, metadata, and a GitHub repository URL.

\begin{tcolorbox}[
  colback=gray!5,
  colframe=gray!50,
  breakable,
  enhanced,
  title={\textbf{Task Construction System Prompt}},
  fonttitle=\bfseries\small,
  left=4pt, right=4pt, top=4pt, bottom=4pt,
  before skip=6pt, after skip=6pt
]
\begin{lstlisting}[
  basicstyle=\ttfamily\scriptsize,
  breaklines=true,
  breakatwhitespace=false,
  columns=fullflexible,
  keepspaces=true,
  breakindent=0pt,
  postbreak={},
  frame=none,
  backgroundcolor=\color{gray!5}
]
# YouTube Tutorial to Task Construction Guide

This guide explains how to convert a single YouTube Godot tutorial (with transcript and GitHub repo) into GameDevBench tasks.

GameDevBench is a multimodal LLM Agent benchmark to test if models can develop games or assist with game development.

## Godot

Godot is installed and usable with the `godot` command.

VERY IMPORTANT: Whenever you run `godot` please ensure you set a timeout of 1 minute.

## Context

You will be working in a single tutorial folder at a time. Each folder contains:

  {data_folder}/{channel_name}/{video_title}/
  +-- transcript.txt      # Full video transcript
  +-- metadata.json       # Video metadata
  +-- github_repo.txt     # GitHub repository URL

### Tips for YouTube Tutorial Processing

1. Transcript Context: Tutorials often explain "why" before "what" - look for action verbs
2. GitHub is Ground Truth: When transcript is unclear, GitHub repo shows what actually works
3. Simplify Complexity: If tutorial covers multiple concepts, break into multiple tasks
4. Test Repository First: Clone and run GitHub repo to understand expected behavior
5. Match Repo Structure: Use similar node names and organization as the repo
6. License Compliance: All repos already filtered for MIT/Apache-2.0/CC0-1.0

### Common Pitfalls

- Copying GitHub Repo Verbatim: Adapt to GameDevBench structure, don't just copy
- Ignoring Transcript: GitHub shows "what" but transcript explains "why" and learning objective
- Overly Broad Tasks: Focus on one specific learning objective per task
- Missing Assets: Ensure sprites/sounds from repo are included in both task directories
- Weak Validation: Check everything that makes the task correct

### Key Principles for Single-Folder Processing

1. All analysis happens in the tutorial folder first
   - Clone repo to repo/ subdirectory
   - Create analysis_progress.md for documentation
   - Complete all analysis before creating tasks
2. Document everything as you go
   - Update analysis_progress.md after each step
   - Include transcript quotes, repo structure, task ideas
   - Track what works and what doesn't
3. Test the GitHub repo before extracting tasks
   - Run godot --import-all --quit
   - Verify it's a working Godot project
   - Check for missing assets or dependencies
4. Navigate to GameDevBench root for task creation
   - Don't create tasks inside the tutorial folder
   - Copy assets from tutorial's repo/ to task directories
5. Return to tutorial folder for final documentation
   - Update analysis_progress.md with completion status
   - Note which tasks were created
   - Record any issues for future reference

## Phase 1: Setup

### Step 1: Check for Godot 4.

We only want to operate on Godot 4 tutorials. If the tutorial folder / github repo is for a Godot 3 project, stop and report that.

### Step 2: Set Up Your Workspace

Create a progress tracking file to document your work.

### Step 3: Read Available Files

Read the transcript, GitHub repo URL, and metadata. Document in analysis_progress.md:
- Main topic, key concepts, estimated complexity
- Has implementation steps: yes/no
- GitHub repo available: yes/no
- Suitable for tasks: yes/no/maybe

### Step 4: Analyze Transcript for Task Ideas

Look for in transcript.txt:
- Node creation mentions ("create a CharacterBody2D")
- Node adjustments ("adjust the anchors of the container")
- Property settings ("set the gravity to 980")
- Script attachment steps ("attach a new script")
- Signal connections ("connect the body_entered signal")
- Scene organization instructions
- Multimodal reasoning ("create an animation from the spritesheet")

Task Categories:
- Graphics & Animation
- Physics & Movement
- World Building
- Programming
- User Interface
- Game Systems
- Audio

IMPORTANT:
- The skillset required between tasks should be diverse. Focus on tasks that require adjusting node properties, adding new nodes, or adding sub-children.
- Focus on keeping the tasks faithful to the tutorial.
- Each task must be independent from each other.
- Tasks that require multimodal reasoning (e.g., cutting up a spritesheet, adjusting sound to match animation) are especially desirable.

### Step 5: Clone and Examine GitHub Repository

Clone to a repo/ subdirectory. Examine structure, key files, project.godot for Godot version. Document:
- Main scene, key scenes, scripts, assets
- Key nodes and configuration
- Critical properties
- Scripts summary

Create a dependency graph mapping file and feature dependencies.

### Step 6: Test the GitHub Repository

Import assets, try to run the project, check for tests. Document import results, project status, and conclusion.

## Phase 2: Task Creation

### Step 1: Extract Actionable Tasks

Based on transcript + GitHub repo, identify specific, testable tasks. Ensure tasks center on node creation and/or inspector configuration.

Difficulty Guidelines:
- Easy: 1 to 3 individual steps
- Medium: 4 to 8 individual steps
- Hard: 9 or more individual steps

For each task document:
- Source (transcript line + repo file)
- What to create (specific nodes/properties)
- Validation criteria
- File modifications from start to finish state
- Difficulty, GitHub reference, categories, multimodal flag

### Step 2: Create Task Directories in GameDevBench

Navigate to GameDevBench root. Determine next task number. Create directories for both tasks/ and tasks_gt/. Copy project template and needed assets from cloned repo.

### Step 3: Create task_config.json

{
  "task_id": XXXX,
  "name": "Descriptive Task Name from Video",
  "instruction": "Clear, specific instructions...",
  "difficulty": "easy|medium|hard",
  "template_id": X,
  "metadata": {
    "tutorial_folder": "...",
    "tutorial_source": "YouTube: {channel} - {video_title}",
    "video_id": "...",
    "github_repo": "...",
    "transcript_excerpt": "...",
    "expected_nodes": ["NodeType1", "NodeType2"],
    "key_properties": {"property": "expected_value"}
  },
  "tags": ["youtube", "2d|3d", "category", "node-type"]
}

### Step 4: Reference GitHub Repo for Ground Truth

- GitHub repo shows the completed task
- Adapt to fit GameDevBench structure (don't copy verbatim)
- Document which files were copied and which were not

### Step 5: Create Ground Truth Implementation

Study GitHub repo scene structure, recreate key nodes and hierarchy, set properties, add scripts, simplify if needed, and test.

## Phase 3: Task Instruction

### Step 1: Create Task Instruction

Key principles:
- Concise, clear, and unambiguous
- Solver must understand requirements to go from start state to ground truth
- Solver will NOT have access to tests or ground truth
- Self-contained: no references to transcript, tests, other tasks, or the tutorial name
- Mention technical requirements (node types, APIs) but not usage details
- NO tips, NO hints, NO test commands, NO code examples

Each instruction step must have evidence pointing back to the original source (transcript or repository).

## Phase 4: Task Validation

### Step 1: Create Validation Script

Create a GDScript validation that:
- Asserts required nodes exist in the correct hierarchy
- Confirms critical inspector values left unset in the starting point
- Fails early when structural requirements are missing
- Prints VALIDATION_PASSED or VALIDATION_FAILED messages
- Copy to BOTH task and ground truth directories

Document how each test maps to a specific instruction step.

### Step 2: Create Starting Point (Incomplete Version)

- Provide basic scaffolding so the scene launches
- Include required raw assets
- Omit key implementation details:
  - Leave tutorial-created nodes absent
  - Skip scripts and signal connections
  - Leave tutorial-modified inspector properties unset

Goal: Starting point should fail validation but provide foundation.

### Step 3: Test and Validate

- Starting point: should output VALIDATION_FAILED
- Ground truth: should output VALIDATION_PASSED

## Final Quality Checklist

- analysis_progress.md fully completed
- Each instruction step lists transcript or repo evidence
- Validation maps to instruction with file + line numbers
- Multiple independent tasks created from the tutorial
- Every task has both ground truth and starting point
- Starting points fail validation
- Ground truths pass validation
- Each task includes valid main.tscn and test.tscn
- Node/inspector-focused requirements asserted
- At least one task requires multimodal reasoning
- Asset transfer summary recorded
- Deviations, blockers, and Godot version issues documented
\end{lstlisting}
\end{tcolorbox}

\section{Task Refinement Prompt}
\label{appendix:task_refinement}

Below is the full prompt provided to the agent for automatic task validation and refinement (Stage 3). The prompt consists of two parts: (1) an instruction that describes the validation workflow and context, and (2) a checklist template that the agent must fill out with evidence for each criterion. If any criterion fails, the agent is instructed to fix the task accordingly.

A variant of this prompt omits scripting-related checks and adds the constraint ``No \texttt{.gd} script editing is required,'' which was used for tasks that focus exclusively on scene construction and inspector configuration.

\begin{tcolorbox}[
  colback=gray!5,
  colframe=gray!50,
  breakable,
  enhanced,
  title={\textbf{Task Refinement System Prompt}},
  fonttitle=\bfseries\small,
  left=4pt, right=4pt, top=4pt, bottom=4pt,
  before skip=6pt, after skip=6pt
]
\begin{lstlisting}[
  basicstyle=\ttfamily\fontsize{7}{8.5}\selectfont,
  breaklines=true,
  breakatwhitespace=false,
  columns=fullflexible,
  keepspaces=true,
  breakindent=0pt,
  postbreak={},
  frame=none,
  aboveskip=0pt,
  belowskip=0pt,
  xleftmargin=0pt,
  xrightmargin=0pt
]
# Benchmark Task Validation Guide

This file documents instructions to validate a task for GameDevBench.

GameDevBench is a multimodal LLM Agent benchmark to test if models can develop games or assist with game development.

## Context

You will be working on a single task at a time. Each task has three related folders:

### Tutorial Folder
  {data_folder}/{channel_name}/{video_title}/
  +-- repo/                   # YouTube tutorial repository
  +-- analysis_progress.md    # Notes from task generation
  +-- transcript.txt          # Full video transcript
  +-- metadata.json           # Video metadata
  +-- github_repo.txt         # GitHub repository URL

### Task (Starting Point) Folder
  tasks/task_{number}_{name}/
  +-- task_config.json        # Contains the task instruction
  +-- scripts/test.gd         # Contains the test code

### Task (Ground Truth) Folder
  tasks_gt/task_{number}_{name}/
  (Same structure as starting point, with completed solution)

## Instructions

Your job is to:
1. Read and analyze the transcript
2. Read and examine the GitHub repository
3. Read and examine the task created
4. Document your progress
5. Validate whether the task satisfies each criterion
5a. Each criterion must have evidence for validation documented

Copy the checklist template into the task starting point folder and fill it out as you validate.

---

# Key Checklist

- [ ] The task starting point runs with `uv run gamedevbench validate $TASK_NAME` and successfully outputs a test failure.
  - Evidence:

- [ ] The task ground truth runs with `uv run gamedevbench --gt validate $TASK_NAME` and successfully outputs SUCCESS.
  - Evidence:

- [ ] In every task, there exists a valid main.tscn and test.tscn similar to tasks_gt/task_0001_place_asset.
  - Evidence:

- [ ] The task instruction matches the tutorial transcript. The task instructions must be a subset of the tutorial transcript.
  - Instruction 1 / Transcript 1
  ...
  - Instruction N / Transcript N
  - Instructions Missing from Transcript: Explanation

- [ ] The task code is directly derived from the repository code. Please document where the derived code is.
  - Evidence:

- [ ] The task instruction is clear, unambiguous, and self-contained. There are no references to the tutorial or other tasks.
  - Evidence:

- [ ] The tests in test.gd match the instructions. All tests are contained in the instruction. Similarly, all instructions are in the tests. Explain how to adjust the tests themselves to match the instructions.
  - Test 1, Instruction 1
  - Test 2, Instruction 2
  ...
  - Test N, Instruction N
  - Missing Coverage: Explanation here

- [ ] Each test in test.gd is unambiguously defined in the instructions. With just the instruction and the task code (without looking at the tests), it is unambiguously possible to satisfy each test condition.
  - For EACH test, list EVERY assertion/check it makes, then verify the instruction specifies that EXACT detail:
  - Test 1, Assertions: [list each check], Instruction coverage: [exact instruction text]
  - Test 2, Assertions: [list each check], Instruction coverage: [exact instruction text]
  ...
  - CRITICAL AMBIGUITY CHECKS - For each test, verify:
    - [ ] String formatting (padding, delimiters, exact format) is specified in instruction
    - [ ] Exact string values/names are in instruction
    - [ ] Number formats (zero-padding, decimal places) are specified
    - [ ] Any comparison operators have clear criteria
    - [ ] Node names, paths, and types match instruction
    - [ ] Property values (numbers, booleans, strings) have exact values in instruction
  - Ambiguous Tests: [List any test checks that lack exact specification in instruction]

- [ ] If there are multiple solutions to the problem, the tests in test.gd are flexible to allow multiple solutions. Mark as completed if there is only one solution and that solution is clearly decipherable from the instructions.
  - Evidence:

- [ ] The folder and file names are consistent with other tasks (tasks_gt/task_0001_place_asset).
  - Evidence:

- [ ] PROCEED. Check this box if the task is validated and all key checks pass successfully.


# Feature Checklist

- [ ] The task contains instructions or goals that are Node/inspector-focused.
  - Evidence:

- [ ] The task contains or requires multimodal reasoning or understanding to complete.
  - Evidence:

- [ ] The task contains a multimodal input (such as an image) in the instruction.
  - Evidence:


# Identifying Ambiguous Tests (Examples)

## Example 1: AMBIGUOUS - Vague formatting requirement

Instruction: "Format incremental and timer step text"

Test Code:
  var expected_text = "Destroy 5 ships 00/05"
  if do_label.text != expected_text:
      issues.append("Initial text should be '%s'" % expected)

Assertions:
- Checks text equals exactly "Destroy 5 ships 00/05"
- Requires zero-padded format
- Requires specific spacing and delimiter

Why AMBIGUOUS: Instruction says "format text" but does not specify zero-padding, exact delimiter, spacing, or format string structure.

How to fix: Change instruction to "Format incremental step text as '{details} {collected:02d}/{required:02d}'" OR make test flexible to accept any reasonable format.

## Example 2: UNAMBIGUOUS - Specific requirement

Instruction: "Set the ColorRect size to Vector2(screen_max_size, screen_max_size)"

Test Code:
  var expected = Vector2(screen_max_size, screen_max_size)
  assert(color_rect.size == expected)

Why UNAMBIGUOUS: Instruction explicitly states the exact Vector2 formula. No ambiguity about what value is expected.

## Example 3: AMBIGUOUS - Missing specific values

Instruction: "Connect to QuestManager signals"

Test Code:
  var required = ["step_updated", "step_complete", "quest_completed", "quest_failed"]
  for signal_name in required:
      if not _has_connection(qm, signal_name, quest_ui):
          issues.append("Must connect to %s" % signal_name)

Why AMBIGUOUS: Instruction says "signals" (vague) but test checks for 4 specific signal names. A solver might connect only 2 signals and technically satisfy "connect to signals".

How to fix: Change instruction to "Connect to QuestManager signals: step_updated, step_complete, quest_completed, and quest_failed".
\end{lstlisting}
\end{tcolorbox}

\section{Human Annotation Instructions}
\label{appendix:human_annotation}

Below are the instructions provided to human annotators during Stage 4. Annotators were asked to verify task correctness, fix common issues, and flag tasks that were unsalvageable.

\definecolor{promptbg}{gray}{0.95}

\begin{tcolorbox}[
  colback=promptbg,
  colframe=gray!60,
  breakable,
  enhanced,
  title={\textbf{Human Annotation Instructions}},
  fonttitle=\bfseries\small,
  left=6pt, right=6pt, top=4pt, bottom=4pt,
  before skip=6pt, after skip=6pt
]
\begin{lstlisting}[
  basicstyle=\ttfamily\fontsize{7}{8.5}\selectfont,
  breaklines=true,
  breakatwhitespace=false,
  columns=fullflexible,
  keepspaces=true,
  breakindent=0pt,
  postbreak={},
  frame=none,
  aboveskip=0pt,
  belowskip=0pt,
  xleftmargin=0pt,
  xrightmargin=0pt
]
# Human Validation

Goal: Ensure that tasks are solvable and not ambiguous. Essentially, we want to ensure that tasks are actually good tasks.

Fixing a task should take 5-15 minutes per task at most. If it takes longer, then it may be too difficult for us to fix, in which case we can skip the task and mark it as Blocked in the Status.

## Common Issues

Most tasks require minor fixes. By far the most common are the following:

- Ambiguous Instructions (e.g., model says to change the skybox, but there may be three different skybox variables possible)
- Overly Strict Tests (e.g., tests require something that isn't stated in the instruction)
  - Naming is a good one (tests for a named node, without requesting it)
- Conflicting Instructions (e.g., model says to do two incompatible things)
- Tutorial References - sometimes the instruction mentions the tutorial; just remove that reference

As with previous, the best way to find these errors is to actually implement the task. However, in this case you should feel free to use any tool (e.g., a coding agent directly) to support you or implement the task. The goal is NOT to see if you can solve the task (as it was before), but to identify errors in the task.

## General Procedure

1. Look at the task (base and ground truth versions) in the editor. Read the task instruction. See if it looks reasonable (multimodality) or if it's clearly a scripting oriented task. What you're looking for is something that just makes sense. Run this before to ensure everything loads properly:
    godot --path /path/to/folder --editor
2. Change directory to the task. Ask your agent of choice to solve the task.
3. While the agent is solving the task, take a look at test.gd. See if each test matches the instruction. If not, fix. You can usually catch some easy errors, such as named node/function tests.
4. After the agent finishes, run validation. See if it passes / fails.
5. If the agent failed, pass the test.gd in. Ask the agent if it missed anything. Ask the agent if the things it missed are due to its own error or due to ambiguous instruction/overly strict tests. This is my usual prompt:
    "Look at test.gd. Was there anything you missed? Did you miss it due to your mistake or were the instructions unclear?"
   The agent is sometimes good, sometimes bad at this. It really depends on the agent.

## Multimodality

You may encounter tasks that have poor multimodality. If it's easy to fix, then just fix it (e.g., something is not in the camera, some default is incorrect). If it's not, flag it and move on. Note that some tasks are inherently multimodal in which case don't worry about it. We have an automated script that is improving multimodality, so really just fix any obvious errors.

## Skipping Tasks

If the project doesn't play, doesn't load, or some other catastrophic issue, just skip the task. There should be very few tasks like this.

Please mark all the issues using the issues dropdown. This is important so we can double back to tasks if necessary.

## Important: Preventing Agent Contamination

When using an agent to attempt the task, always run the task in the task folder itself and instruct the model:
- Don't look at test.gd, test.tscn or execute any tests
- Don't look outside of your current folder
- Don't look at task_config.json
- Don't look at task_validation.md
\end{lstlisting}
\end{tcolorbox}


\newpage
\section{Prompt Templates}
\label{sec:prompt_templates}

All task prompts are derived from the same base instruction, with optional extensions that provide additional multimodal feedback mechanisms.
We report the exact prompt templates used for the baseline, MCP-enabled, and runtime-video-enabled settings below.

\begin{tcolorbox}[
  colback=promptbg,
  colframe=gray!60,
  breakable,
  enhanced,
  title={\textbf{Baseline Prompt}},
  fonttitle=\bfseries\small,
  left=6pt, right=6pt, top=4pt, bottom=4pt,
  before skip=6pt, after skip=6pt
]
\begin{lstlisting}[
  basicstyle=\ttfamily\fontsize{7}{8.5}\selectfont,
  breaklines=true,
  columns=fullflexible,
  keepspaces=true,
  frame=none
]
<INSTRUCTION FROM task_config.json>

You must complete the full task without any further assistance.
Godot is installed and you can run godot using the `godot` command.
It is recommended to run this with a timeout (e.g., `timeout 10 godot`)
to prevent hanging.
You are a visual agent and can use images and videos to help you
understand the state of the game.
\end{lstlisting}
\end{tcolorbox}

\begin{tcolorbox}[
  colback=promptbg,
  colframe=gray!60,
  breakable,
  enhanced,
  title={\textbf{Prompt with MCP}},
  fonttitle=\bfseries\small,
  left=6pt, right=6pt, top=4pt, bottom=4pt,
  before skip=6pt, after skip=6pt
]
\begin{lstlisting}[
  basicstyle=\ttfamily\fontsize{7}{8.5}\selectfont,
  breaklines=true,
  columns=fullflexible,
  keepspaces=true,
  frame=none
]
<INSTRUCTION FROM task_config.json>

You must complete the full task without any further assistance.
Godot is installed and you can run godot using the `godot` command.
It is recommended to run this with a timeout (e.g., `timeout 10 godot`)
to prevent hanging.
You are a visual agent and can use images and videos to help you
understand the state of the game.

You have access to a Godot MCP (Model Context Protocol) server that
provides specialized tools for working with Godot projects.

Available MCP Tools:
- `godot-screenshot`: Takes a screenshot of the Godot editor.

Usage Guidelines:
- Use screenshots before starting to understand project structure.
- Use screenshots after making changes to verify correctness.
- Use screenshots during debugging to inspect editor state.
\end{lstlisting}
\end{tcolorbox}

\begin{tcolorbox}[
  colback=promptbg,
  colframe=gray!60,
  breakable,
  enhanced,
  title={\textbf{Prompt with Runtime Video}},
  fonttitle=\bfseries\small,
  left=6pt, right=6pt, top=4pt, bottom=4pt,
  before skip=6pt, after skip=6pt
]
\begin{lstlisting}[
  basicstyle=\ttfamily\fontsize{7}{8.5}\selectfont,
  breaklines=true,
  columns=fullflexible,
  keepspaces=true,
  frame=none
]
<INSTRUCTION FROM task_config.json>

You must complete the full task without any further assistance.
Godot is installed and you can run godot using the `godot` command.
It is recommended to run this with a timeout (e.g., `timeout 10 godot`)
to prevent hanging.
You are a visual agent and can use images and videos to help you
understand the state of the game.

You can run the game and capture visual output using:
- `godot --path . --quit-after 1 --write-movie output.png`

You can capture short videos using:
- `timeout 60s godot --path . --quit-after 60 --write-movie output.avi`

Ensure that Godot exits after execution to avoid hanging.
Use images or videos to verify that your changes worked as expected.
\end{lstlisting}
\end{tcolorbox}

%
%
%
%
%
%
%

\newpage
\section{Task Examples}

We provide examples of tasks in \texttt{GameDevBench}.
Each task can be solved by taking actions in the editor as a human would or by directly editing code files.

\label{appendix:task-example}
\subsection{Isometric Crusader Animation }

In this example, the goal is to add physical collision
and animation to the character.
This is a \textbf{2D graphics and animations} task that focuses on the animation editor which is a \textbf{contextual editor}.
\begin{figure}[h]
    \centering
    \includegraphics[width=1\linewidth]{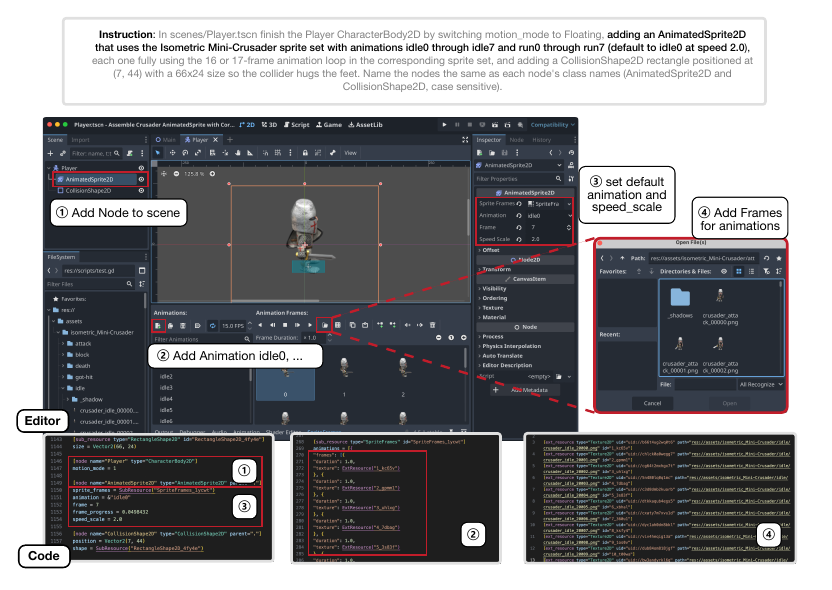}
    \caption{An example task from \texttt{GameDevBench}. In this example, the goal is to add physical collision and animation to the character.
    This can be achieved through either taking actions directly in the editor or editing code files. 
    Each action in the editor is equivalent to specific modifications within the code files.
    Matching steps are denoted with the same numbers in our figure.
    }
    \label{fig:example_crusader}
\end{figure}
\clearpage
\subsection{Floating Balls}

In this example, the goal is to populate an empty 3D scene with a water depth visualization, including environment lighting, shader-driven water plane, background spheres, and a camera.
This is a \textbf{3D graphics and animations} task that focuses on the \textbf{scene editor}.

\begin{figure}[h!]
    \centering
    \includegraphics[width=1.0\linewidth]{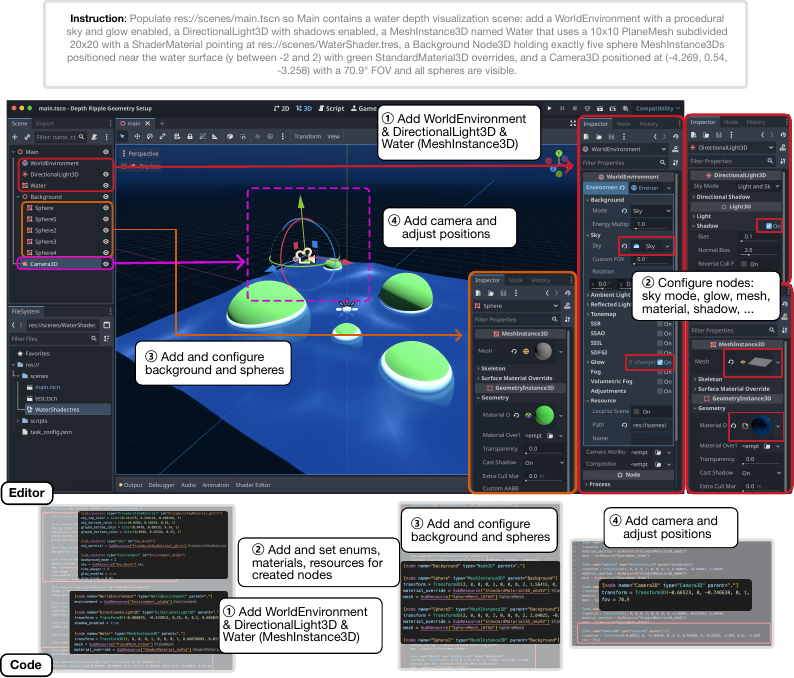}
    \caption{An example task from \texttt{GameDevBench}. In this example, the goal is to            populate an empty 3D scene with a water depth visualization, including environment lighting, shader-driven water plane, background spheres, and a camera.
      This can be achieved through either taking actions directly in the
      editor or editing the scene file (\texttt{main.tscn}). Each action in the editor
      is equivalent to specific modifications within the scene file. Matching
      steps are denoted with the same numbers in our figure
    }
    \label{fig:example_3d}
\end{figure}
\clearpage
\subsection{FPS User Interface}

In this example, the goal is to build a complete three-screen menu system (Launch, Pause, and Restart) and signal connections to the menu handler script.
This is a \textbf{user interface} task that focuses on the \textbf{scene editor}.

\begin{figure}[h]
    \centering
    \includegraphics[width=0.95\linewidth]{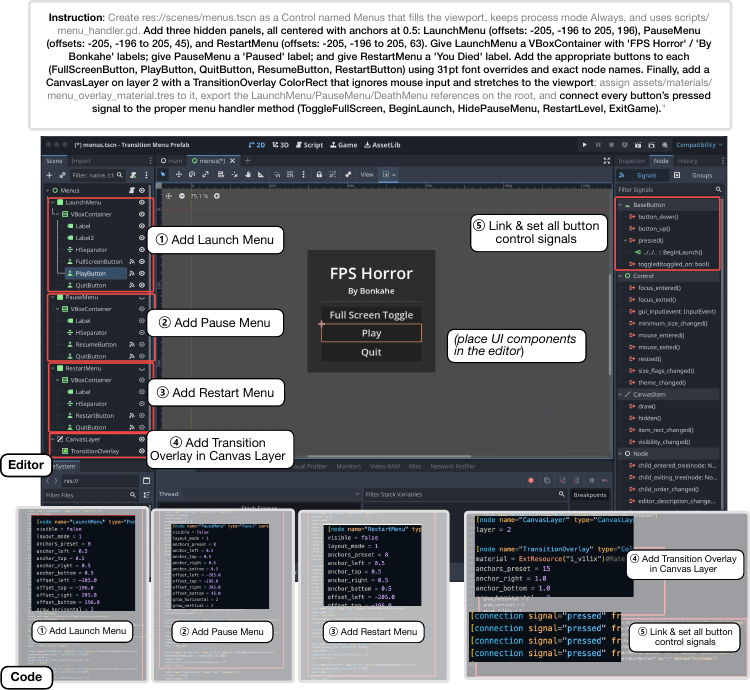}
    \caption{An example task from \texttt{GameDevBench}. In this example, the goal is to build a complete three-screen menu system (Launch, Pause, and Restart) with styled buttons, title labels, a shader-driven transition overlay, and signal connections to the menu handler script. This can be achieved through either taking actions directly in the editor or editing the scene file (\texttt{menus.tscn}). Each action in the editor is equivalent to specific modifications within the scene file. Matching steps are denoted with the same numbers in our figure.
    }
    \label{fig:example_interface}
\end{figure}
\clearpage
\subsection{RTS Unit}
In this example, the goal is to build a reusable
RTS unit with a sprite, collision shapes, a detection area for neighbor avoidance, and an aura shader
that highlights the unit when selected. 
The main focus is on the scripting, thus this is a \textbf{gameplay logic} task that focuses on the \textbf{script editor}.

\begin{figure}[h]
    \centering
    \includegraphics[width=0.95\linewidth]{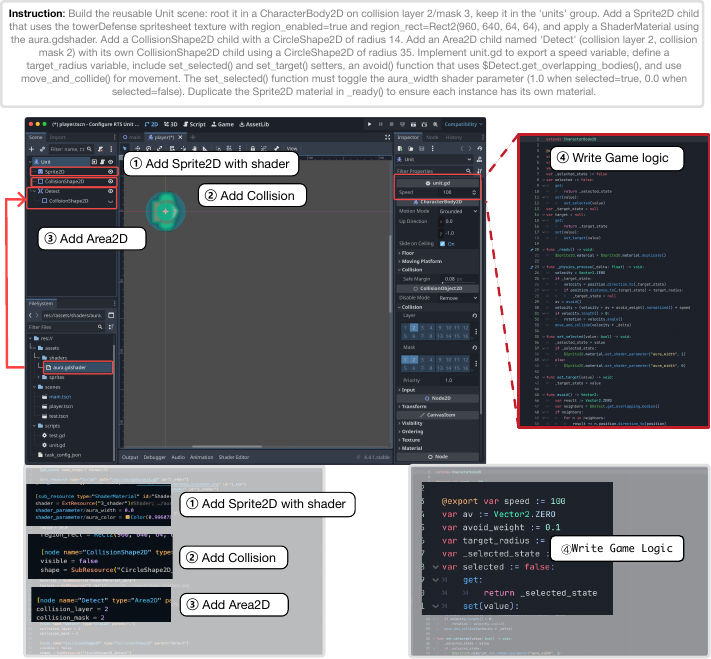}
    \caption{An example task from \texttt{GameDevBench}. In this example, the goal is to build a reusable RTS unit with a sprite, collision shapes, a detection area for neighbor avoidance, and an aura shader that highlights the unit when selected. Unlike purely scene-based tasks, this task requires both editing the scene file (\texttt{player.tscn}) and implementing gameplay logic in a script file (\texttt{unit.gd}). Each action in the editor is equivalent to specific modifications within the code files. Matching steps are denoted with the same numbers in our figure.
    } \label{fig:example_gamelogic}
\end{figure}



\newpage
\section{Task Statistics}
\label{appendix:statistics}

We provide detailed statistics for \texttt{GameDevBench}. Different tasks test different skills resulting in drastically different data distributions.
For example, some sprite animation tasks require thousands of sprites to be processed.

\begin{table*}[h!]
\centering
\caption{Comprehensive GameDevBench Task Statistics. Mean ($3\sigma$) denotes the mean after excluding values more than 3 standard deviations from the mean.}
\label{tab:gamedevbench-stats-comprehensive}
\begin{tabular}{llrrrr}
\toprule
 & & \textbf{Mean} & \textbf{Mean ($3\sigma$)} & \textbf{Median} & \textbf{Max} \\
\midrule
\multirow{4}{*}{Overview} & Files & 67.7 & 18.7 & 10.0 & 1929 \\
 & Filetypes & 6.7 & 6.6 & 6.0 & 18 \\
 & Lines of Code & 518.9 & 392.7 & 204.0 & 20072 \\
 & Nodes & 19.7 & 13.5 & 7.0 & 982 \\
\midrule
\multirow{2}{*}{Godot Scripting} & Scripting Files & 3.0 & 2.4 & 2.0 & 49 \\
 & Scripting Lines & 236.7 & 176.6 & 107.5 & 9543 \\
\midrule
\multirow{2}{*}{Godot Scenes} & Scene Files & 3.6 & 3.2 & 3.0 & 54 \\
 & Scene Lines & 219.5 & 154.6 & 32.5 & 10282 \\
\midrule
\multirow{5}{*}{Assets} & Images & 53.4 & 4.3 & 1.0 & 1920 \\
 & Image Size (px) & 119.5K & 73.3K & 71.8K & 16.8M \\
 & Shaders & 0.2 & 0.1 & 0.0 & 7 \\
 & Audio & 1.0 & 0.2 & 0.0 & 13 \\
 & Resources & 0.7 & 0.4 & 0.0 & 14 \\
\midrule
\multirow{6}{*}{Gold Patch} & Files Edited & 4.7 & 4.5 & 4.0 & 17 \\
 & Filetypes Edited & 3.2 & 3.2 & 3.0 & 6 \\
 & Total Lines Edited & 114.1 & 70.2 & 48.5 & 1949 \\
 & Scripting Lines Edited & 14.9 & 11.3 & 0.0 & 208 \\
 & Scene Lines Edited & 92.2 & 47.5 & 24.0 & 1949 \\
 & Nodes Edited & 2.3 & 1.9 & 1.0 & 24 \\
\bottomrule
\end{tabular}
\end{table*}

\newpage
\section{Case Study of Model Failure}\label{appendix:case_study}
\subsection{Common Game Development Patterns}\label{appendix:case_study:common}

\begin{figure}[h]
    \centering
    \includegraphics[width=\linewidth]{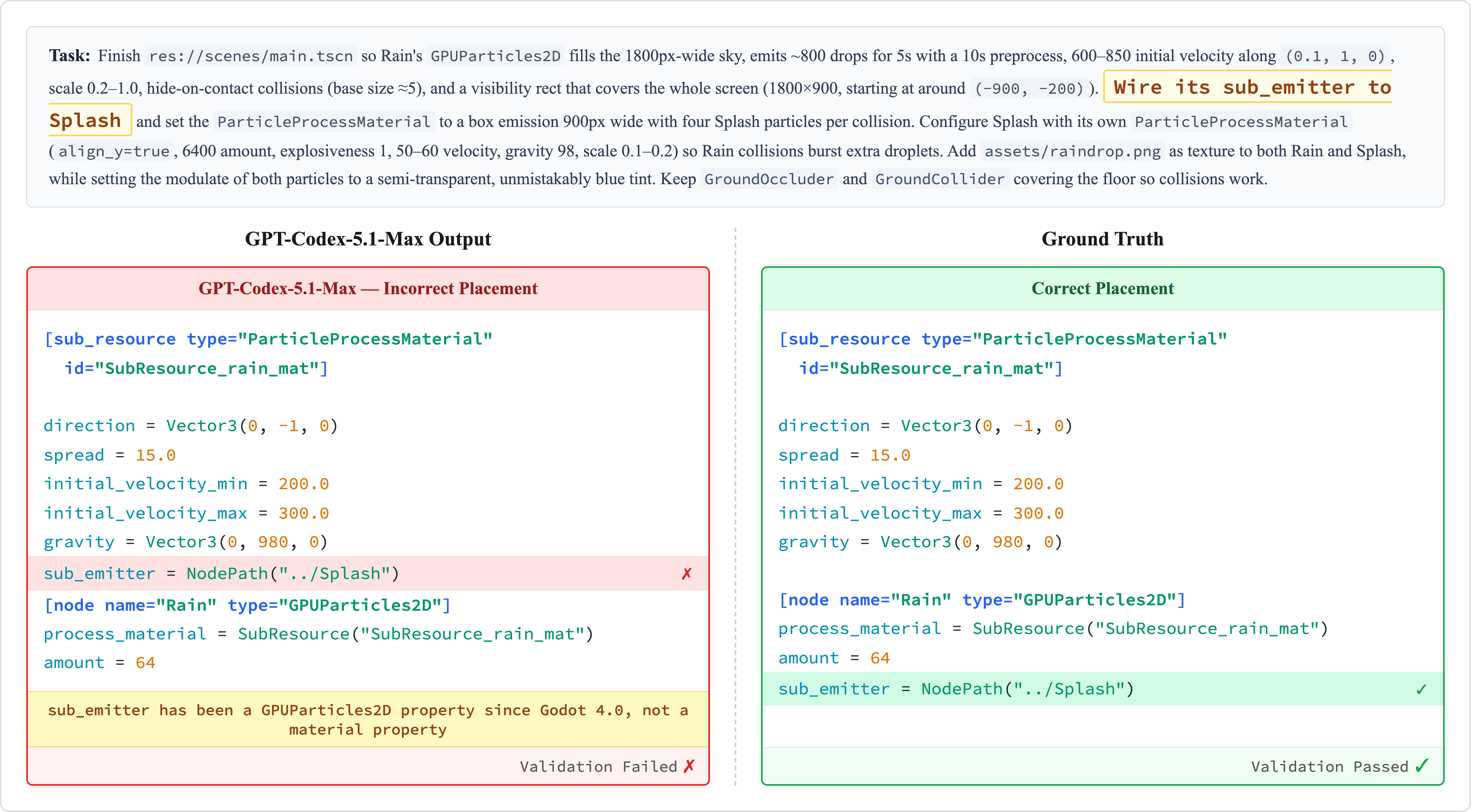}
    \caption{Example of Godot common game development task. \model{GPT-5.4} places \texttt{sub\_emitter} inside the \texttt{ParticleProcessMaterial} sub-resource (left, red) instead of on the \texttt{GPUParticles2D} node (right, green). The property belongs to \texttt{GPUParticles2D}.}
    \label{fig:common_case_study}
\end{figure}

Figure~\ref{fig:common_case_study} shows a representative failure of common game development. The task requires completing a Godot \texttt{.tscn} scene file for a rain particle system, including wiring the \texttt{sub\_emitter} property on a \texttt{GPUParticles2D} node to a sibling \texttt{Splash} node. \model{GPT-5.4} produces the correct property name and value (\texttt{sub\_emitter = NodePath("../Splash")}), but places it under the \texttt{ParticleProcessMaterial} sub-resource instead of the \texttt{GPUParticles2D} node. The \texttt{sub\_emitter} property belongs to \texttt{GPUParticles2D} and has no meaning on a material resource, indicating that the model lacks the knowledge that this property must be placed under the \texttt{GPUParticles2D} node. 

\newpage
%
%
%

%
\section{Detailed Benchmark Results}
\label{appendix:details}
\subsection{Full Benchmark Result with Token Count and Cost}
Table~\ref{tab:task_results_detail} reports detailed benchmark results for both the core task set and the full task set(containing variant tasks deriving from core tasks). Overall, stronger models generally achieve higher pass@1, and multimodal inputs provide gains in performance across models. We also include average token usage and cost to provide a reference for comparing effectiveness and efficiency across agent and model setups.

\begin{table*}[h]
  \centering
  \caption{Core tasks indicates the tasks we build from the pipeline in Section \ref{sec:benchmark-construction}. 
  Full tasks also include task variations derived from core tasks. 
  Tokens and costs are per-task medians over all records contributing to each row (k = 1{,}000
  tokens).
  Task variants were designed to explicitly test for multimodal understanding.
  Full \texttt{pass@1} is lower due to this increased requirement for multimodal understanding.
  Trends in general hold, except \model{GPT-5.4} and \model{Gemini 3 Pro} swap first and second place.
  }
  \label{tab:task_results_detail}
  \renewcommand{\arraystretch}{1.05}
  \setlength{\tabcolsep}{3pt}
  \scriptsize
  \resizebox{0.98\textwidth}{!}{%
  \begin{tabular}{@{}l l c c c c r r@{}}
  \toprule
  Harness & Model & w/ Screenshot & w/ Video & Core pass@1 (\%) & Full pass@1 (\%) & Median Tokens & Median Cost (\$) \\
  \midrule
  \multirow{6}{*}{\centering \framework{claude-code}}
  & \multirow{4}{*}{\model{claude-haiku-4-5-20251001}}
  & \textcolor{red}{\texttimes} & \textcolor{red}{\texttimes} & 20.8 & 13.8 & 10.5k & 0.170 \\
  & & \textcolor{green}{\checkmark} & \textcolor{red}{\texttimes} & 21.4 & 15.6 & 5.7k & 0.193 \\
  & & \textcolor{red}{\texttimes} & \textcolor{green}{\checkmark} & 25.3 & 18.6 & 11.6k & 0.223 \\
  & & \textcolor{green}{\checkmark} & \textcolor{green}{\checkmark} & 24.7 & 16.5 & 10.3k & 0.188 \\
  \cline{2-8}
  & \multirow{2}{*}{\model{claude-sonnet-4-5-20250929}}
  & \textcolor{red}{\texttimes} & \textcolor{red}{\texttimes} & 31.2 & 28.8 & 6.6k & 0.237 \\
  & & \textcolor{green}{\checkmark} & \textcolor{green}{\checkmark} & 41.6 & 34.8 & 8.2k & 0.395 \\
  \midrule

  \multirow{6}{*}{\centering \framework{codex}}
  & \multirow{4}{*}{\model{gpt-5.4-mini}}
  & \textcolor{red}{\texttimes} & \textcolor{red}{\texttimes} & 48.1 & 36.9 & 329k & 0.279 \\
  & & \textcolor{green}{\checkmark} & \textcolor{red}{\texttimes} & 46.1 & 37.8 & 546k & 0.418 \\
  & & \textcolor{red}{\texttimes} & \textcolor{green}{\checkmark} & 52.6 & 43.2 & 848k & 0.612 \\
  & & \textcolor{green}{\checkmark} & \textcolor{green}{\checkmark} & 48.7 & 39.0 & 786k & 0.581 \\
  \cline{2-8}
  & \multirow{2}{*}{\model{gpt-5.4}}
  & \textcolor{red}{\texttimes} & \textcolor{red}{\texttimes} & 49.4 & 41.1 & 188k & 0.156 \\
  & & \textcolor{green}{\checkmark} & \textcolor{green}{\checkmark} & \textbf{56.5} & \emph{52.0} & 865k & 0.522 \\
  \midrule

  \multirow{6}{*}{\centering \framework{gemini-cli}}
  & \multirow{4}{*}{\model{gemini-3-flash-preview}}
  & \textcolor{red}{\texttimes} & \textcolor{red}{\texttimes} & 47.4 & 45.4 & 296k & 0.056 \\
  & & \textcolor{green}{\checkmark} & \textcolor{red}{\texttimes} & 48.1 & 45.4 & 284k & 0.058 \\
  & & \textcolor{red}{\texttimes} & \textcolor{green}{\checkmark} & 51.9 & 46.9 & 464k & 0.082 \\
  & & \textcolor{green}{\checkmark} & \textcolor{green}{\checkmark} & 50.0 & 44.1 & 582k & 0.098 \\
  \cline{2-8}
  & \multirow{2}{*}{\model{gemini-3-pro-preview}}
  & \textcolor{red}{\texttimes} & \textcolor{red}{\texttimes} & 48.7 & 50.1 & 329k & 0.199 \\
  & & \textcolor{green}{\checkmark} & \textcolor{green}{\checkmark} & \emph{54.5} & \textbf{53.8} & 492k & 0.265 \\
  \midrule

  \multirow{10}{*}{\centering \framework{openhands}}
  & \multirow{2}{*}{\model{claude-haiku-4-5-20251001}}
  & \textcolor{red}{\texttimes} & \textcolor{red}{\texttimes} & 22.7 & 15.6 & 1{,}433k & 0.354 \\
  & & \textcolor{green}{\checkmark} & \textcolor{green}{\checkmark} & 27.9 & 17.7 & 1{,}853k & 0.459 \\
  \cline{2-8}
  & \multirow{2}{*}{\model{gemini-3-flash-preview}}
  & \textcolor{red}{\texttimes} & \textcolor{red}{\texttimes} & 31.8 & 30.3 & 350k & 0.119 \\
  & & \textcolor{green}{\checkmark} & \textcolor{green}{\checkmark} & 36.4 & 31.8 & 583k & 0.181 \\
  \cline{2-8}
  & \multirow{2}{*}{\model{gpt-5.4-mini}}
  & \textcolor{red}{\texttimes} & \textcolor{red}{\texttimes} & 44.8 & 38.4 & 809k & 0.243 \\
  & & \textcolor{green}{\checkmark} & \textcolor{green}{\checkmark} & 44.8 & 36.9 & 1{,}182k & 0.299 \\
  \cline{2-8}
  & \multirow{2}{*}{\model{kimi-k2.5}}
  & \textcolor{red}{\texttimes} & \textcolor{red}{\texttimes} & 31.2 & 18.9 & 245k & 0.070 \\
  & & \textcolor{green}{\checkmark} & \textcolor{green}{\checkmark} & 31.2 & 20.7 & 487k & 0.103 \\
  \cline{2-8}
& \multirow{2}{*}{\model{qwen3.5-397b}}
  & \textcolor{red}{\texttimes} & \textcolor{red}{\texttimes} & 9.7 & 5.4 & 145k & 0.499 \\
  & & \textcolor{green}{\checkmark} & \textcolor{green}{\checkmark} & 7.1 & 5.1 & 215k & 0.716 \\
  \bottomrule
  \end{tabular}%
  }
\end{table*}

\newpage
\subsection{Benchmark Result Based on Easy/Difficult Task Split}

We annotate tasks into \texttt{easy} and \texttt{hard} subsets. 
We find that almost every model configuration performs worse on the hard subset.
The only exception is \model{Gemini 3 Pro} which actually \textit{improves} in the harder subset.
We hypothesize this is due to the model's strong inherent multimodal understanding.

\begin{table*}[h]
  \centering
  \caption{\texttt{pass@1} (\%) split by task difficulty. $\Delta$ is the percentage-point difference (\texttt{hard} $-$ \texttt{easy}).}
  \label{tab:partition_hard_easy}
  \renewcommand{\arraystretch}{1.05}
  \setlength{\tabcolsep}{5pt}
  \begin{tabular}{l l c c c c c}
  \toprule
  Harness & Model & w/ Screenshot & w/ Video & Easy (\%) & Hard (\%) & $\Delta$ (pp) \\
  \midrule
  \multirow{6}{*}{\centering \framework{claude-code}} & \multirow{4}{*}{\model{claude-haiku-4-5-20251001}} & \textcolor{red}{\texttimes} & \textcolor{red}{\texttimes} &
  22.1 & 8.4 & \textcolor{red}{$-$13.7} \\
   &  & \textcolor{green}{\checkmark} & \textcolor{red}{\texttimes} & 21.4 & 11.9 & \textcolor{red}{$-$9.5} \\
   &  & \textcolor{red}{\texttimes} & \textcolor{green}{\checkmark} & 26.0 & 13.9 & \textcolor{red}{$-$12.1} \\
   &  & \textcolor{green}{\checkmark} & \textcolor{green}{\checkmark} & 24.4 & 11.4 & \textcolor{red}{$-$13.0} \\
  \cline{2-7}
   & \multirow{2}{*}{\model{claude-sonnet-4-5-20250929}} & \textcolor{red}{\texttimes} & \textcolor{red}{\texttimes} & 33.6 & 25.7 & \textcolor{red}{$-$7.8} \\
   &  & \textcolor{green}{\checkmark} & \textcolor{green}{\checkmark} & 46.6 & 27.2 & \textcolor{red}{$-$19.3} \\
  \midrule

  \multirow{6}{*}{\centering \framework{codex}}
   & \multirow{4}{*}{\model{gpt-5.4-mini}} & \textcolor{red}{\texttimes} & \textcolor{red}{\texttimes} & 51.9 & 27.2 & \textcolor{red}{$-$24.7} \\
   &  & \textcolor{green}{\checkmark} & \textcolor{red}{\texttimes} & 48.9 & 30.7 & \textcolor{red}{$-$18.2} \\
   &  & \textcolor{red}{\texttimes} & \textcolor{green}{\checkmark} & 56.5 & 34.7 & \textcolor{red}{$-$21.8} \\
   &  & \textcolor{green}{\checkmark} & \textcolor{green}{\checkmark} & 55.0 & 28.7 & \textcolor{red}{$-$26.2} \\
  \cline{2-7}
  & \multirow{2}{*}{\model{gpt-5.4}} & \textcolor{red}{\texttimes} & \textcolor{red}{\texttimes} & 51.1 & 34.7 & \textcolor{red}{$-$16.5} \\
   &  & \textcolor{green}{\checkmark} & \textcolor{green}{\checkmark} & 61.1 & 46.0 & \textcolor{red}{$-$15.0} \\
  \midrule

  \multirow{6}{*}{\centering \framework{gemini-cli}} & \multirow{4}{*}{\model{gemini-3-flash-preview}} & \textcolor{red}{\texttimes} & \textcolor{red}{\texttimes} & 50.4
  & 42.1 & \textcolor{red}{$-$8.3} \\
   &  & \textcolor{green}{\checkmark} & \textcolor{red}{\texttimes} & 51.9 & 41.1 & \textcolor{red}{$-$10.8} \\
   &  & \textcolor{red}{\texttimes} & \textcolor{green}{\checkmark} & 48.9 & 45.5 & \textcolor{red}{$-$3.3} \\
   &  & \textcolor{green}{\checkmark} & \textcolor{green}{\checkmark} & 52.7 & 38.6 & \textcolor{red}{$-$14.1} \\
  \cline{2-7}
   & \multirow{2}{*}{\model{gemini-3-pro-preview}} & \textcolor{red}{\texttimes} & \textcolor{red}{\texttimes} & 46.6 & 52.5 & \textcolor{green}{$+$5.9} \\
   &  & \textcolor{green}{\checkmark} & \textcolor{green}{\checkmark} & 52.7 & 54.5 & \textcolor{green}{$+$1.8} \\
  \midrule

  \multirow{10}{*}{\centering \framework{openhands}} & \multirow{2}{*}{\model{claude-haiku-4-5-20251001}} & \textcolor{red}{\texttimes} & \textcolor{red}{\texttimes} &
  23.7 & 10.4 & \textcolor{red}{$-$13.3} \\
   &  & \textcolor{green}{\checkmark} & \textcolor{green}{\checkmark} & 31.3 & 8.9 & \textcolor{red}{$-$22.4} \\
  \cline{2-7}
   & \multirow{2}{*}{\model{gemini-3-flash-preview}} & \textcolor{red}{\texttimes} & \textcolor{red}{\texttimes} & 35.1 & 27.2 & \textcolor{red}{$-$7.9} \\
   &  & \textcolor{green}{\checkmark} & \textcolor{green}{\checkmark} & 40.5 & 26.2 & \textcolor{red}{$-$14.2} \\
  \cline{2-7}
   & \multirow{2}{*}{\model{gpt-5.4-mini}} & \textcolor{red}{\texttimes} & \textcolor{red}{\texttimes} & 49.6 & 31.2 & \textcolor{red}{$-$18.4} \\
   &  & \textcolor{green}{\checkmark} & \textcolor{green}{\checkmark} & 50.4 & 28.2 & \textcolor{red}{$-$22.2} \\
  \cline{2-7}
   & \multirow{2}{*}{\model{kimi-k2.5}} & \textcolor{red}{\texttimes} & \textcolor{red}{\texttimes} & 33.6 & 9.4 & \textcolor{red}{$-$24.2} \\
   &  & \textcolor{green}{\checkmark} & \textcolor{green}{\checkmark} & 36.6 & 10.4 & \textcolor{red}{$-$26.2} \\
  \cline{2-7}
   & \multirow{2}{*}{\model{qwen3.5-397b}} & \textcolor{red}{\texttimes} & \textcolor{red}{\texttimes} & 11.5 & 1.5 & \textcolor{red}{$-$10.0} \\
   &  & \textcolor{green}{\checkmark} & \textcolor{green}{\checkmark} & 8.4 & 3.0 & \textcolor{red}{$-$5.4} \\
  \bottomrule
  \end{tabular}
\end{table*}

\newpage
\section{Failure Analysis}
  \label{app:failure_analysis}

  We analyze failures from the four best-performing model configurations using an LLM-as-a-judge procedure. Consistent with the main paper, failures most often involve
  incorrect game-development patterns or failures of multimodal grounding. Since a single failed task may exhibit multiple error modes, percentages are computed over
  total failures and are not mutually exclusive.

  \begin{table}[h]
  \centering
  \caption{Common game-development failure patterns.}
  \label{tab:game_dev_failures}
  \setlength{\tabcolsep}{6pt}
  \renewcommand{\arraystretch}{1.05}
  \begin{tabular}{l r}
  \toprule
  Failure Type & Failures (\%) \\
  \midrule
  Missing or mis-parented nodes & 63.4 \\
  Missing method, property, or custom signal & 36.2 \\
  Unset exported reference & 35.9 \\
  Incorrect physics setup & 32.1 \\
  Incorrect scene or asset instantiation & 28.2 \\
  Incorrect UI control-tree structure & 25.1 \\
  Incorrect node type & 22.6 \\
  Incorrect TileMap structure & 1.7 \\
  \bottomrule
  \end{tabular}
  \end{table}

  \begin{table}[h]
  \centering
  \caption{Common multimodal-understanding failure patterns.}
  \label{tab:multimodal_failures}
  \setlength{\tabcolsep}{6pt}
  \renewcommand{\arraystretch}{1.05}
  \begin{tabular}{l r}
  \toprule
  Failure Type & Failures (\%) \\
  \midrule
  Incorrect shader or material assignment & 22.6 \\
  Incorrect shader, post-processing, or environment parameters & 22.6 \\
  Incorrect UI layout, spacing, sizing, or anchoring & 19.9 \\
  Incorrect animation state, direction, or frame sequence & 17.8 \\
  Incorrect camera framing, position, or view transform & 17.1 \\
  Incorrect spritesheet region or atlas slice & 15.3 \\
  Incorrect texture, tile, or visual asset selection & 15.3 \\
  Incorrect particle emission, spread, or lifetime parameters & 15.0 \\
  Collider placement inconsistent with visible sprite geometry & 2.4 \\
  \bottomrule
  \end{tabular}
  \end{table}

\section{Data Contamination Analysis}
  \label{app:data_leakage}

  To check for any data contamination, we evaluate whether models can reproduce held-out tutorial content from partial context. For each of 117 tutorial transcripts, we provide the first half of the
  transcript and ask the model to complete the second half. We then compare the generated continuation against the true held-out continuation using ROUGE-L and BLEU-4. A
  score of 1 would indicate perfect memorization. Both models obtain very low overlap scores, suggesting no evidence of systemic memorization.

  \begin{table}[h]
  \centering
  \caption{Transcript continuation overlap on 117 held-out tutorial transcripts.}
  \label{tab:memorization_analysis}
  \setlength{\tabcolsep}{8pt}
  \renewcommand{\arraystretch}{1.05}
  \begin{tabular}{l c c}
  \toprule
  Model & ROUGE-L & BLEU-4 \\
  \midrule
  Gemini 3 Flash & 0.0615 & 0.0172 \\
  GPT 5.1 & 0.0623 & 0.0173 \\
  \bottomrule
  \end{tabular}
  \end{table}

\end{document}